\pdfoutput=1
\documentclass[10pt, logo, onecolumn, copyright]{nv}

\definecolor{linkc}{rgb}{0, 0.44, 0.74}
\definecolor{eqc}{rgb}{1, 0, 0}
\definecolor{newcitecolor}{rgb}{0,0.6,0}
\usepackage{url}
\usepackage{booktabs}
\usepackage{wrapfig}
\usepackage{xspace}
\usepackage{textgreek}
\usepackage{bm}
\usepackage{multirow}
\usepackage[dvipsnames]{xcolor}
\usepackage{graphicx}
\usepackage{collectbox}
\usepackage{adjustbox}
\usepackage{colortbl}
\usepackage{orcidlink}
\usepackage{footmisc}
\usepackage{algorithm}
\usepackage{algpseudocode}
\usepackage{amsmath}
\usepackage{amsfonts}
\usepackage{listings}
\usepackage{caption}
\usepackage{enumitem}
\usepackage{csquotes}
\usepackage{marvosym}
\usepackage{mdframed}
\usepackage[utf8]{inputenc} %
\usepackage[T1]{fontenc}    %

\usepackage{nicefrac}       %
\usepackage{microtype}      %
\usepackage{multicol}
\usepackage{tabto}

\usepackage{dblfloatfix}
\usepackage{times}
\usepackage{verbatim}
\usepackage{amssymb}
\usepackage{mathtools}
\usepackage{subcaption}
\usepackage{array}
\usepackage{bbm}
\usepackage{makecell}
\usepackage{siunitx}
\usepackage{pifont}
\usepackage{pdflscape}
\usepackage{tabularx}
\usepackage{arydshln}
\usepackage{hhline}
\usepackage{diagbox}
\usepackage{tcolorbox}
\usepackage[square,sort,comma,numbers]{natbib}
\usepackage{hyperref}
\usepackage{cleveref}
\usepackage{fp} %
\usepackage{authblk}

\DeclareRobustCommand{\iconhref}[3]{%
  \texorpdfstring{%
    \href{#1}{%
      \raisebox{-0.15em}{\includegraphics[height=1em]{#2}}\,#3%
    }%
  }{#3}%
}

\definecolor{mygreen}{RGB}{34,139,34}
\definecolor{mylightblue}{RGB}{0,162,230}
\definecolor{deepyellow}{RGB}{255, 215, 0} 
\definecolor{catgray}{gray}{0.92}
\definecolor{nvidiagreen}{HTML}{76B900}
\definecolor{codebg}{RGB}{245, 245, 245} 
\definecolor{keywordcolor}{RGB}{0, 0, 153} 
\definecolor{commentcolor}{RGB}{34, 139, 34} 
\definecolor{stringcolor}{RGB}{163, 21, 21}
\definecolor{numbercolor}{RGB}{128, 128, 128}

\makeatletter
\def\blfootnote#1{\xdef\@thefnmark{}\@footnotetext{\scriptsize #1}}
\makeatother

\hypersetup{
    colorlinks=true,
    urlcolor=nvidiagreen,
}

\definecolor{darkred}{rgb}{0.459,0.0,0.08}
\definecolor{myblue}{HTML}{47B1E1}
\definecolor{mygreen}{HTML}{46D45A}
\definecolor{myorange}{HTML}{F3AA84}
\definecolor{usercolor}{HTML}{B41601}       
\definecolor{assistantcolor}{HTML}{76B900}  
\definecolor{linkc}{rgb}{0,0.44,0.74}
\definecolor{eqc}{rgb}{1,0,0}
\definecolor{newcitecolor}{rgb}{0,0.6,0}
\newcommand{\modelname}{SANA-WM\xspace}

\newcommand{\keycell}[1]{\cellcolor{assistantcolor!10}#1}      %
\newcommand{\rankfirst}[1]{\cellcolor{assistantcolor!32}#1}
\newcommand{\ranksecond}[1]{\cellcolor{assistantcolor!20}#1}
\newcommand{\rankthird}[1]{\cellcolor{assistantcolor!10}#1}

\let\cite\citep

\title{\modelname: Efficient Minute-Scale World Modeling with Hybrid Linear Diffusion Transformer}

\author{%
\vspace{-1.5em}
\centering
\fontsize{10pt}{18pt}\selectfont
Haoyi Zhu\textsuperscript{$*$}, ~~ Haozhe Liu\textsuperscript{$*$}, ~~ Yuyang Zhao\textsuperscript{$*$}, ~~ Tian Ye\textsuperscript{$*$}, ~~ Junsong Chen\textsuperscript{$*$}, ~~ Jincheng Yu ~~ 
\\
\vspace{0.4em}
\textbf{\fontsize{10pt}{18pt}\selectfont
Tong He, ~~ Song Han, ~~ Enze Xie ~~ 
}
\\
\vspace{2.5mm}
{\normalsize NVIDIA} \\ %
\vspace{0.3em}
{\normalsize $^*$Equal contribution.} \\
\vspace{0.3em}
{\normalsize
\iconhref{https://nvlabs.github.io/Sana/WM/}{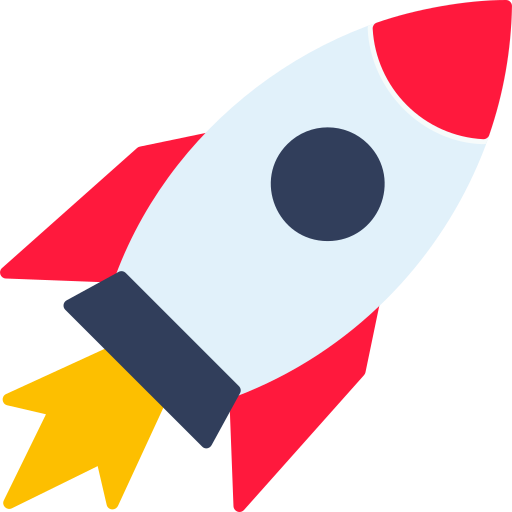}{Project Page}
\quad
\iconhref{https://github.com/NVlabs/Sana}{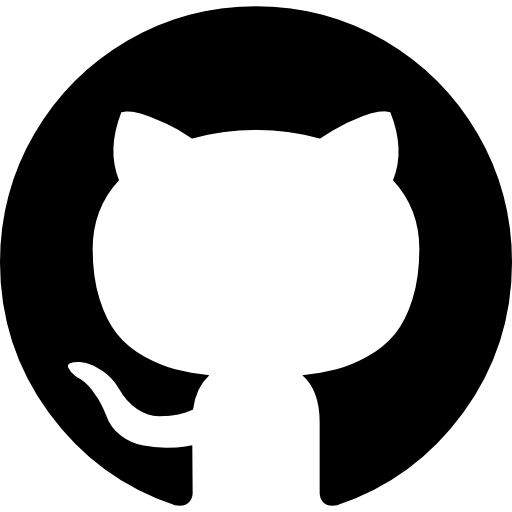}{GitHub}
}
}

\begin{abstract}
\noindent \textbf{Abstract:} We introduce \modelname, an efficient 2.6B-parameter open-source world model natively trained for one-minute generation, synthesizing high-fidelity, 720p, minute-scale videos with precise camera control. \modelname achieves visual quality comparable to large-scale industrial baselines such as LingBot-World and HY-WorldPlay, while significantly improving efficiency. Four core designs drive our architecture: (1) \textbf{Hybrid Linear Attention} combines frame-wise Gated DeltaNet (GDN) with softmax attention for memory-efficient long-context modeling. (2) \textbf{Dual-Branch Camera Control} ensures precise 6-DoF trajectory adherence. (3) \textbf{Two-Stage Generation Pipeline} applies a long-video refiner to stage-1 outputs, improving quality and consistency across sequences. (4) \textbf{Robust Annotation Pipeline} extracts accurate metric-scale 6-DoF camera poses from public videos to yield high-quality, spatiotemporally consistent action labels. Driven by these designs, \modelname demonstrates remarkable efficiency across data, training compute, and inference hardware: it uses only $\sim$213K public video clips with metric-scale pose supervision, completes training in 15 days on 64 H100s, and generates each 60s clip on a single GPU; its distilled variant can be deployed on a single RTX 5090 with NVFP4 quantization to denoise a \textbf{60s 720p clip in 34s}. On our one-minute world-model benchmark, \modelname demonstrates stronger action-following accuracy than prior open-source baselines and achieves comparable visual quality at $36\times$ higher throughput for scalable world modeling.
\end{abstract}

\begin{document}
\maketitle

\begin{figure}[H]
  \centering
  \includegraphics[width=\textwidth]{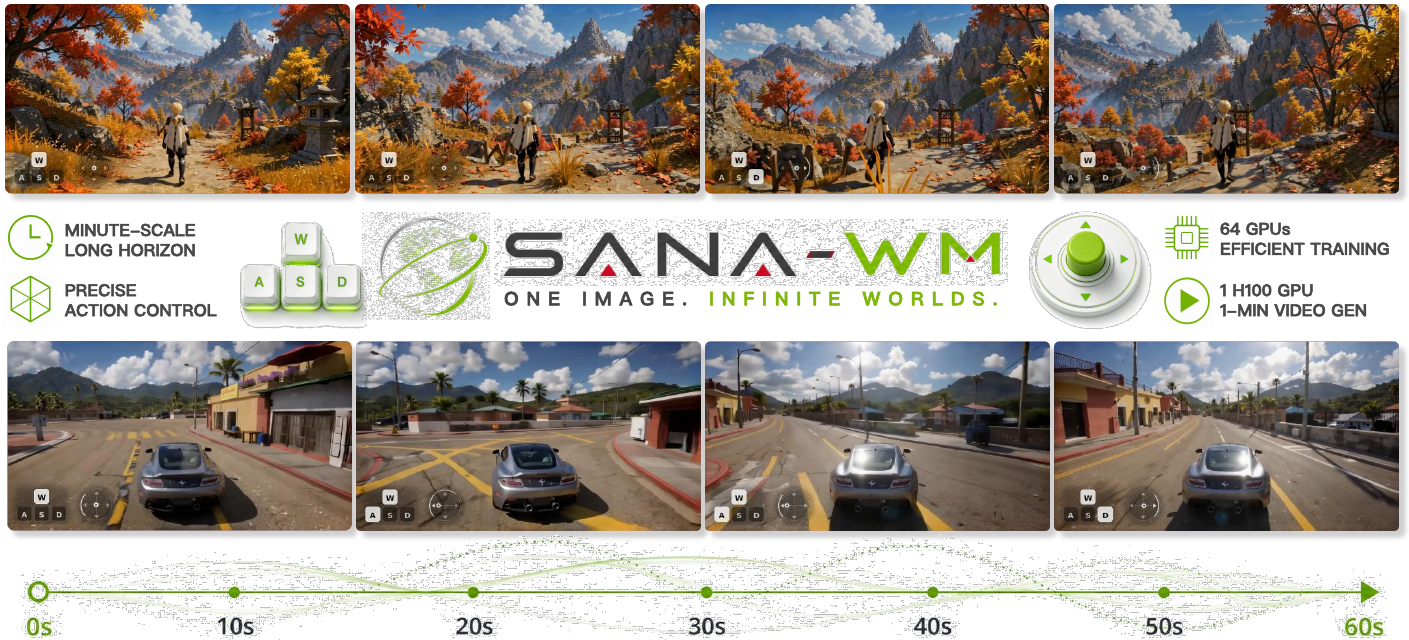}
  \caption{\textbf{\modelname teaser.}
    From one image and an action trajectory, \modelname generates minute-scale 720p worlds with precise control, 64-GPU training, and single-GPU inference.
    }
  \label{fig:teaser}
\end{figure}

\begin{abstract}

We introduce \modelname, an efficient 2.6B-parameter open-source world model natively trained for one-minute generation, synthesizing high-fidelity, 720p, minute-scale videos with precise camera control. \modelname achieves visual quality comparable to large-scale industrial baselines such as LingBot-World and HY-WorldPlay, while significantly improving efficiency. Four core designs drive our architecture: (1) \textbf{Hybrid Linear Attention} combines frame-wise Gated DeltaNet (GDN) with softmax attention for memory-efficient long-context modeling. (2) \textbf{Dual-Branch Camera Control} ensures precise 6-DoF trajectory adherence. (3) \textbf{Two-Stage Generation Pipeline} applies a long-video refiner to stage-1 outputs, improving quality and consistency across sequences. (4) \textbf{Robust Annotation Pipeline} extracts accurate metric-scale 6-DoF camera poses from public videos to yield high-quality, spatiotemporally consistent action labels. Driven by these designs, \modelname demonstrates remarkable efficiency across data, training compute, and inference hardware: it uses only $\sim$213K public video clips with metric-scale pose supervision, completes training in 15 days on 64 H100s, and generates each 60s clip on a single GPU; its distilled variant can be deployed on a single RTX 5090 with NVFP4 quantization to denoise a \textbf{60s 720p clip in 34s}. On our one-minute world-model benchmark, \modelname demonstrates stronger action-following accuracy than prior open-source baselines and achieves comparable visual quality at $36\times$ higher throughput for scalable world modeling.

\end{abstract}

\section{Introduction}
\label{sec:introduction}

World models are becoming a key interface for embodied simulation and interactive environments~\citep{world_models,genie3,gaia1,dreamdojo,aether,hy_worldplay,lingbot_world,infinite_world}.
We study camera-controlled world modeling: given a first frame, text, and a 6-DoF camera trajectory, the model synthesizes a one-minute 720p video that follows the input motion while preserving scene identity.
Recent open-source systems achieve minute-scale, action-conditioned rollouts~\citep{infinite_world,matrixgame3,lingbot_world,hy_worldplay}, but typically require large models, large-scale data, long training schedules, and multi-GPU inference.
A tempting lower-cost alternative is to distill long-rollout models from short-video generators, but such short-horizon teachers provide limited supervision for minute-scale scene persistence and trajectory following.
We therefore ask: \textit{can we natively train a high-fidelity, camera-controllable, one-minute world model while keeping data, training, and inference costs accessible?}

\begin{figure}[t]
  \centering
  \includegraphics[width=0.92\linewidth]{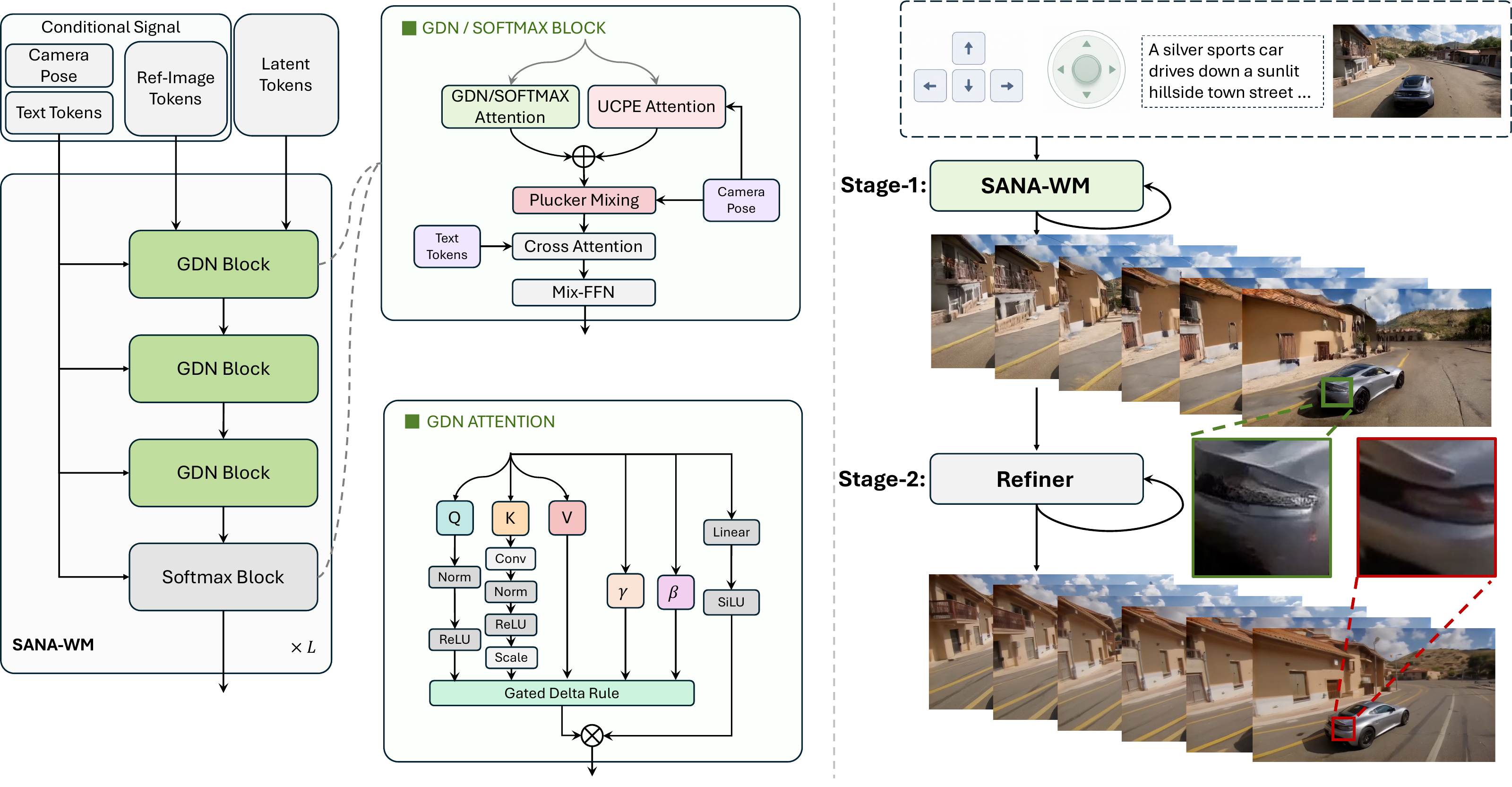}
  \caption{\textbf{\modelname Architecture.} 
  Text, video, and pose tokens pass through alternating GDN and softmax attention blocks. Geometry-aware components (UCPE attention and Pl\"ucker mixing) are integrated to enable pose-conditioned generation, followed by a refiner to improve visual quality.
  }
  \label{fig:pipeline_overview}
\end{figure}

\noindent We introduce \modelname, a 2.6B-parameter open-source video world model designed around efficiency as a first-class objective.
\modelname is trained for one-minute generation using only ${\sim}213$K public video clips with metric-scale pose supervision and 15 days on 64 H100 GPUs.
At inference time, it supports three single-GPU variants: a bidirectional generator for high-quality offline synthesis, a chunk-causal autoregressive generator for sequential rollout, and a few-step distilled autoregressive generator for faster deployment.
Fig.~\ref{fig:teaser} shows representative generations, and Fig.~\ref{fig:pipeline_overview} summarizes the training and inference pipeline.
The improvements mainly lie in four key components.

\noindent\textbf{Efficient Native One-Minute Backbone.}
One-minute 720p generation stresses both token count and long-context modeling, so \modelname pairs a high-compression LTX2 tokenizer~\citep{ltx2} with a hybrid Linear DiT backbone.
The backbone combines frame-wise Gated DeltaNet~\citep{gated_deltanet} blocks for efficient recurrent context aggregation with periodic softmax attention for exact long-range recall.
This design keeps minute-scale context affordable while preserving the modeling capacity needed for scene persistence and camera-conditioned motion.

\noindent\textbf{Dual-Branch Camera Control.}
Precise action-conditioned world modeling requires generated videos to faithfully follow continuous action trajectories, rather than merely aligning with text prompts.
\modelname therefore uses a dual-rate camera conditioning design: a latent-rate UCPE branch~\citep{ucpe} captures global trajectory structure, while a raw-frame Pl\"ucker mixing branch restores fine camera motion inside each temporal VAE stride.
This lets the model preserve control accuracy despite aggressive video compression.

\noindent\textbf{Two-Stage Visual Refinement.}
To further improve visual quality, \modelname adopts a two-stage generation pipeline with a dedicated refinement stage.
We adapt an independent refiner
to operate on long \modelname outputs, correcting structural artifacts and sharpening details across the full minute.
This refinement stage is used as a quality-improvement pass after stage-1 generation.

\noindent\textbf{Robust Data Annotation and Evaluation Benchmark.}
To train camera-controlled videos without proprietary action labels, we build a robust annotation pipeline that recovers accurate metric-scale camera poses from public videos using pose and geometry estimators~\citep{vipe,pi3x,moge2}.
After filtering, this pipeline yields ${\sim}213$K video clips with precise metric-scale pose annotation.
Since existing benchmarks do not target minute-scale world modeling, we build a one-minute benchmark for action following, visual quality, and efficiency.
It contains 80 initial scenes generated by Nano Banana Pro~\citep{google2025nanobananapro} across four scene types, each paired with two revisit trajectories.

\noindent Experiments show that \modelname achieves higher accuracy in action-following than prior open-source baselines, with comparable visual quality, while delivering up to $36\times$ higher generation throughput.
Most importantly for accessibility, it reduces minute-scale generation to a single-GPU inference setting: the bidirectional and chunk-causal variants fit within one H100, and our distilled variant brings 1-minute video generation to 34s on a single RTX 5090 with NVFP4 quantization.

\noindent In summary, our contributions are: (i) a natively one-minute-trained, 720p, action-controllable world model with accessible training and inference cost; (ii) an efficiency-oriented architecture combining high-compression video latents, hybrid GDN/softmax long-context modeling, and dual-branch camera control; (iii) a long-video second-stage refiner that improves stage-1 visual quality; and (iv) a robust data annotation and evaluation pipeline for long-horizon world modeling.

\section{Related Work}
\label{sec:related_work}

\noindent\textbf{Long-video generation and interactive world models.}
Large video generators increasingly use diffusion or flow-transformer backbones over compressed spatiotemporal latents, with representative systems including Stable Video Diffusion, Sora, CogVideoX, Wan, HunyuanVideo, MovieGen, Cosmos, LTX-Video/LTX2, and SANA-Video~\citep{stable_video_diffusion,sora,cogvideox,wan,hunyuanvideo,moviegen,cosmos,ltx_video,ltx2,chen2025sana}.
Long-duration generation is commonly approached through autoregressive or block-wise rollout, diffusion forcing, streaming training, and memory- or cache-aware inference~\citep{skyreels_v2,magi,huang2024selfforcing,longlive,chen2025sana}.
World-model research spans several related but distinct settings: latent predictive models for control and planning~\citep{world_models,dreamerv3}, representation-centric predictive models that learn visual or video abstractions without directly generating pixels~\citep{i_jepa,v_jepa,v_jepa2}, and generative simulators that roll out observations under actions or conditions~\citep{gaia1,genie,genie2,genie3,gamengen,oasis,gamegen,matrixgame,matrixgame3,relic,live_world,astra,magicworld,lingbot_world,infinite_world,yume,hy_worldplay}.
Interactive world models extend video generation toward action-conditioned simulation, supporting keyboard, gamepad, camera, text, robot, or mixed controls over long rollouts~\citep{dreamzero,dreamdojo,genie3,gamegen,matrixgame,matrixgame3,unisim,yume,aether,hy_worldplay,lingbot_world,worldcam,ucm,captain_safari,deepverse}.
A parallel line studies explicit memory, scene persistence, and geometry-aware state for revisits and long-horizon consistency, including BEV or occupancy-based driving simulators, camera-aware memories, reconstruction-based methods, and spatially persistent 3D/4D world-generation systems~\citep{drivedreamer,driveworld,ucm,mosaicmem,captain_safari,memorize_when_needed,vggt_world,aether,deepverse,versecrafter,hy_world2,inspatio_world,worldcam}.

\noindent\textbf{Camera control, geometry, and action spaces.}
Action-conditioned world models differ substantially in their control interface.
Some systems use robot or embodied actions~\citep{dreamzero,dreamdojo}, some use keyboard or gamepad controls for games~\citep{genie3,gamegen,matrixgame,matrixgame3}, and others use language, events, or mixed high-level commands~\citep{unisim,yume}.
Camera-controlled generation is closely related to novel-view synthesis and geometric video generation: CameraCtrl and MotionCtrl add camera-control modules to pretrained video diffusion models, CamCo combines Pl\"ucker conditioning with epipolar constraints, and ViewCrafter and SEVA use generative view synthesis to produce target-camera video from one or more input views~\citep{cameractrl,motionctrl,camco,viewcrafter,seva}.
Camera representations include raw extrinsics and intrinsics, epipolar or geometric constraints, dense Pl\"ucker raymaps~\citep{plucker_lfn}, and relative or unified camera positional encodings such as UCPE~\citep{ucpe,li2025cameras}.
Longer camera-controlled rollouts further benefit from camera-pose memory, spatial warping, or persistent 3D/4D scene representations~\citep{aether,ucm,worldcam,captain_safari,deepverse,versecrafter,hy_world2,inspatio_world}.
Pose and depth recovery methods such as VIPE, Pi3/Pi3X, MoGe-2, VGGT, and WinT3R are complementary tools for estimating metric geometry from public videos or generated rollouts rather than video generators themselves~\citep{vipe,pi3x,moge2,vggt,wint3r}.

\noindent\textbf{Efficient sequence models for long visual horizons.}
Standard softmax attention remains effective and can be accelerated by kernels such as FlashAttention~\citep{flash_attention2}, but its memory and compute grow with context length.
Efficient long-context modeling has therefore moved beyond pure softmax attention toward linear attention, kernelized attention, gated linear attention, state-space models, convolutional mixers, test-time-training layers, and delta-rule recurrences~\citep{linear_attention,performer,gla,rwkv,retnet,hyena,mamba,mamba2,ttt_layers,gated_deltanet}.
Recent long-context language architectures combine recurrent, linear, or state-space layers with occasional exact-attention or sparse modules to recover selected long-range information while keeping most layers efficient~\citep{qwen3,qwen3_next,kimi_linear,kimi_k2}.
Beyond language modeling, these efficient mechanisms are also entering visual generation: SANA and SANA-Video use linear-attention backbones for image and video diffusion generation, while high-compression tokenizers such as DC-AE, DC-VideoGen, and LTX-style VAEs reduce the number of visual tokens processed by the generator~\citep{sana,chen2025sana,dc_ae,dc-video-gen,ltx_video,ltx2}.

\noindent\textbf{Data, annotation, and metrics.}
Camera-controllable world modeling depends on data with reliable geometry, diverse motion, and long-horizon scene coverage.
Existing sources include internet video datasets, real-estate and spatial-video collections, 3D captures, embodied-scene datasets, game or synthetic environments, and image-generation pipelines for controlled benchmark construction~\citep{realestate10k,dl3dv,spatialvid,miradata,omniworld,sekai,google2025nanobananapro}.
For filtering and enhancement, prior work uses shot detection, video quality assessment, optical flow, 3D Gaussian reconstruction, and diffusion-based restoration tools~\citep{transnetv2,dover,unimatch,fcgs,difix3d}.
Evaluation commonly combines perceptual video quality, learned perceptual similarity, generated-video distribution metrics, and recovered-camera trajectory accuracy~\citep{vbench,lpips,unterthiner2018towards,pi3x,umeyama}.

\section{Method}
\label{sec:method}

\modelname is a world model designed to generate minute-long high-resolution videos with precise camera control under strict efficiency constraints. 
Scaling to this regime introduces three key challenges: (i) the prohibitive compute and memory cost of modeling 720p sequences; (ii) accurate, high-frequency action conditioning on continuous 6-DoF camera trajectories; and (iii) degraded visual quality when a base generator is trained under limited data and compute. 
To address these, \modelname builds on SANA-Video~\citep{chen2025sana} with three complementary designs: a hybrid GDN/softmax attention architecture for efficient long-context modeling, dual-rate camera conditioning for coarse-to-fine trajectory control, and a second-stage refiner for minute-length video to improve fidelity.

\subsection{Progressive Training Strategy}
\label{sec:training_strategy}

We train progressively from short clips to minute-scale videos, increasing sequence length and introducing architectural components in four stages:

\noindent\textbf{Stage 1: Efficient VAE Adaptation.}
To make minute-scale video modeling computationally feasible, we replace the baseline VAE~\citep{wan, dc-video-gen} with LTX2-VAE~\citep{ltx2} to leverage its superior spatiotemporal compression ratio. 
Given changed channel dimensions, we discard the original patchify layer and final output projection, and re-initialize them from scratch to align with the LTX2 space. 
Full-model fine-tuning adapts the model to this new latent distribution in 50k steps. 
The representation is $2.0\times$ smaller than ST-DC-AE and $8.0\times$ smaller than Wan2.1-VAE, improving training and inference efficiency.

\noindent\textbf{Stage 2: Hybrid Architecture Adaptation.} 
To improve the efficiency--quality trade-off of the backbone, we adapt the pre-trained SANA-Video model to the Hybrid GDN-Softmax architecture (Sec.~\ref{sec:long_context_modeling}). This architecture change is first optimized on short video clips, where training is cheaper and failure modes are easier to diagnose, before scaling to longer sequences.

\noindent\textbf{Stage 3: Minute-Scale Extension and Action Conditioning.} 
After stabilizing the architecture, we extend the sequence length to minute-scale videos to enable long-horizon temporal modeling. In parallel, we incorporate Dual-Branch Camera Control (Sec.~\ref{sec:camera_control}) to support metric 6-DoF trajectory conditioning, allowing explicit control over camera motion.

\noindent\textbf{Stage 4: Chunk-Causal Fine-Tuning and Few-Step Distillation.}
Starting from the one-minute bidirectional camera-control model, we further fine-tune a chunk-causal variant for autoregressive rollout.
We then use self-forcing distillation~\citep{huang2024selfforcing} to reduce sampling to four denoising steps.
For deployment, we add attention-sink tokens and local temporal windows to the softmax attention layers, keeping softmax memory and per-chunk latency constant with respect to rollout length.

\subsection{Memory-Efficient Long-Context Modeling}
\label{sec:long_context_modeling}

As background, SANA-Video~\citep{chen2025sana} uses ReLU-based cumulative linear attention in place of causal softmax attention.
For latent frame $t\ge0$ with $S=H_\ell W_\ell$ spatial tokens, let $\mathbf{Q}_t,\mathbf{K}_t,\mathbf{V}_t\in\mathbb{R}^{D\times S}$ collect the per-head queries, keys, and values, and let $\phi(\cdot)=\mathrm{ReLU}(\cdot)$.
Unlike softmax attention, which forms pairwise weights from $\mathbf{Q}_t^\top\mathbf{K}_{\le t}$, cumulative linear attention accumulates key--value outer products before applying the current queries:
{\small
\begin{equation}
\label{eq:cumulative_linear_attention}
\widetilde{\mathbf{O}}^{\mathrm{LA}}_t
=
\underbrace{\left(\sum_{\tau=0}^{t}\mathbf{V}_\tau\phi(\mathbf{K}_\tau)^\top\right)}_{\mathbf{A}^{\mathrm{LA}}_t}
\phi(\mathbf{Q}_t)
=
\left(\mathbf{A}^{\mathrm{LA}}_{t-1}+\mathbf{V}_t\phi(\mathbf{K}_t)^\top\right)\phi(\mathbf{Q}_t).
\end{equation}
}
Eq.~\ref{eq:cumulative_linear_attention} shows only the unnormalized numerator for brevity; the standard linear-attention denominator is omitted.
With $\mathbf{A}^{\mathrm{LA}}_{-1}=\mathbf{0}$, the cumulative numerator state $\mathbf{A}^{\mathrm{LA}}_t\in\mathbb{R}^{D\times D}$ is updated once per latent frame after aggregating all spatial-token outer products, so memory stays constant.

\noindent \textbf{Limitations of Cumulative Linear Attention.}
This compact state has no explicit decay or saliency mechanism: stale features accumulate with the same effective weight as more recent ones. At the minute scale, the unbounded growing state causes drift and degrades training stability.

\noindent \textbf{From Token-wise GDN to Frame-wise GDN.}
Gated DeltaNet (GDN)~\citep{gated_deltanet} augments the same recurrent state with a decay gate and a delta-rule correction:
\begin{equation}
\label{eq:token_gdn}
\mathbf{S}_i
=\gamma_i\mathbf{S}_{i-1}
+\big(\mathbf{v}_i-\gamma_i\mathbf{S}_{i-1}\hat{\mathbf{k}}_i\big)\beta_i\hat{\mathbf{k}}_i^\top,
\qquad
\mathbf{o}_i=\mathbf{S}_i\hat{\mathbf{q}}_i .
\end{equation}
Here $\mathbf{S}_i\in\mathbb{R}^{D\times D}$ is the token-wise recurrent state, $\hat{\mathbf{q}}_i,\hat{\mathbf{k}}_i,\mathbf{v}_i\in\mathbb{R}^{D}$ are the normalized query, normalized key, and value vectors, $\beta_i\in[0,1]$ is an update gate, and $\gamma_i\in(0,1]$ is a decay gate.
The correction term updates only the residual between the target value and the current state prediction, while $\gamma_i$ forgets obsolete content.
Standard GDN scans one token per recurrent step.
Our video model instead scans one latent frame per step.
For frame $t$, let $\hat{\mathbf{Q}}_t,\hat{\mathbf{K}}_t,\mathbf{V}_t\in\mathbb{R}^{D\times S}$ collect the query features, key features, and values used by frame-wise GDN; our additional key scaling is defined in the stabilization step below.
Let $\gamma_t\in(0,1]$ be a frame-level decay, and let $\boldsymbol{\beta}_t=\mathrm{diag}(\beta_{t,1},\ldots,\beta_{t,S})$ be per-token update gates.
The frame-wise state update becomes
\begin{equation}
\label{eq:frame_gdn}
\mathbf{S}_t
=\mathbf{S}_{t-1}\mathbf{M}_t+\mathbf{U}_t,\qquad
\mathbf{M}_t=\gamma_t\!\left(\mathbf{I}-\hat{\mathbf{K}}_t\boldsymbol{\beta}_t\hat{\mathbf{K}}_t^\top\right),\qquad
\mathbf{U}_t=\mathbf{V}_t\boldsymbol{\beta}_t\hat{\mathbf{K}}_t^\top,\qquad
\mathbf{O}_t=\mathbf{S}_t\hat{\mathbf{Q}}_t .
\end{equation}
Here $\mathbf{S}_t,\mathbf{M}_t,\mathbf{U}_t\in\mathbb{R}^{D\times D}$ are frame recurrent state, transition matrix, and additive update, respectively, and $\mathbf{O}_t\in\mathbb{R}^{D\times S}$ contains the output tokens for frame $t$.
Thus the recurrent state remains $D\times D$, while one recurrent step consumes all $S$ spatial tokens from a latent frame.

\noindent \textbf{Algebraic Stabilization for Spatial Explosion.}
Since $\mathbf{S}_t$ is repeatedly multiplied by $\mathbf{M}_t$, the transition should be non-expansive.
Let $\bar{\mathbf{K}}_t=[\bar{\mathbf{k}}_{t,s}]_{s=1}^{S}=\mathrm{ReLU}(\mathrm{RMSNorm}(\mathbf{K}_t))$ and $\mathbf{A}_t=\bar{\mathbf{K}}_t\boldsymbol{\beta}_t\bar{\mathbf{K}}_t^\top$.
The unscaled key energy is
\begin{equation}
\label{eq:gdn_key_mass}
\operatorname{tr}(\mathbf{A}_t)
=\sum_{s=1}^{S}\beta_{t,s}\|\bar{\mathbf{k}}_{t,s}\|_2^2 ,
\end{equation}
Since $\mathbf{A}_t$ is positive semidefinite, $\lambda_{\max}(\mathbf{A}_t)\le\operatorname{tr}(\mathbf{A}_t)$; an $O(S)$ trace can make $\mathbf{I}-\mathbf{A}_t$ expansive.
We therefore scale only the keys:
\begin{equation}
\label{eq:k_scale}
\hat{\mathbf{K}}_t = \bar{\mathbf{K}}_t \cdot \frac{1}{\sqrt{D \cdot S}} .
\end{equation}
With RMS-normalized keys and $\beta_{t,s}\in[0,1]$, $\operatorname{tr}(\hat{\mathbf{K}}_t\boldsymbol{\beta}_t\hat{\mathbf{K}}_t^\top)\le1$, hence $\|\mathbf{M}_t\|_2\le\gamma_t\le1$; $1/\sqrt{D}$ matches token-wise GDN L2 key normalization, and the extra $1/\sqrt{S}$ averages over spatial tokens.

\noindent \textbf{Bidirectional and Chunk-Causal GDN Variants.}
We use the same recurrence bidirectionally by summing forward and reversed-time scans. For chunk-causal inference, we partition latent frames into chunks, keep the forward scan global, and reset the reversed scan at chunk boundaries, giving each chunk local future context without leakage.

\noindent \textbf{Hybrid GDN/Softmax Attention.}
To enhance long-video generation performance, we further fine-tune the GDN model by replacing every fourth block with standard softmax attention~\citep{flash_attention2}, while retaining the original QKV and output projections.

\subsection{Dual-Branch Camera Control}
\label{sec:camera_control}

We use dual-rate geometric conditioning: latent-frame UCPE~\citep{ucpe} captures global 6-DoF pose, while raw-frame Pl\"ucker mixing compensates motion inside each VAE stride.

\noindent \textbf{Coarse Branch: Ray-Local UCPE.}
For each latent token at frame $t$ and spatial cell $s$, let $\mathbf{T}^{\mathrm{c2w}}_t=[\mathbf{R}_t\,|\,\mathbf{o}_t]$ be the camera-to-world pose and let $\mathbf{A}_t$ be the camera intrinsic matrix.
We unproject the corresponding pixel with $\mathbf{A}_t$ and transform it by $\mathbf{T}^{\mathrm{c2w}}_t$, obtaining a world-space ray with camera center $\mathbf{o}_{t}\in\mathbb{R}^{3}$ and unit direction $\mathbf{d}_{t,s}\in\mathbb{R}^{3}$.
We build a ray-local basis $\mathbf{z}=\operatorname{norm}(\mathbf{d}_{t,s})$, $\mathbf{x}=\operatorname{norm}(\mathbf{u}_{t}\times\mathbf{z})$, and $\mathbf{y}=\mathbf{z}\times\mathbf{x}$, where $\mathbf{u}_{t}$ is the camera vertical axis and $\operatorname{norm}$ denotes $L_2$ normalization.
This defines a homogeneous ray transform $\mathbf{D}_{t,s}\in\mathbb{R}^{4\times4}$ from world coordinates to the ray-local frame.
Following UCPE~\citep{ucpe}, we split each camera-branch attention-head vector into geometric channels and standard RoPE channels.
For token $i=(t,s)$, define $\mathbf{D}_i=\mathbf{D}_{t,s}$ and let $\mathrm{RoPE}_i$ be the standard spatiotemporal rotary operator for token $i$.
We apply the ray-local transform to the geometric channels and RoPE to the remaining channels:
\begin{equation}
\begin{aligned}
\widetilde{\mathbf{Q}}^{c}_{i} &=
  \big(\mathbf{D}_{i}^{\top}\oplus \mathrm{RoPE}_{i}\big)\mathbf{Q}^{c}_{i},\\
(\widetilde{\mathbf{K}}^{c}_{i},\widetilde{\mathbf{V}}^{c}_{i}) &=
  \big(\mathbf{D}_{i}^{-1}\oplus \mathrm{RoPE}_{i}\big)(\mathbf{K}^{c}_{i},\mathbf{V}^{c}_{i}),\\
\mathbf{O}^{c}_{i} &=
  \big(\mathbf{D}_{i}\oplus \mathrm{RoPE}_{i}^{-1}\big)
  \mathrm{GDN}_{\mathrm{cam}}(\widetilde{\mathbf{Q}}^{c},\widetilde{\mathbf{K}}^{c},\widetilde{\mathbf{V}}^{c})_{i},
\end{aligned}
\label{eq:ucpe_camera_branch}
\end{equation}
where superscript $c$ denotes the camera branch, $\mathbf{Q}^c_i,\mathbf{K}^c_i,\mathbf{V}^c_i\in\mathbb{R}^{D}$ are per-head camera-branch query/key/value vectors, and $\widetilde{\cdot}$ denotes the pose-transformed representation.
The operator $\oplus$ denotes a block-diagonal composition over the UCPE and RoPE channel groups; $\mathbf{D}_i$ is applied blockwise to 4-D homogeneous coordinate groups within the UCPE channels.
The camera branch uses its own QKV projections but shares the frame-wise GDN gates with the main branch; its zero-initialized projection is added to the main attention output.

\noindent \textbf{Fine Branch: Raw-Frame Pl\"ucker Mixing.}
The coarse UCPE branch operates at the latent-frame rate, whereas each latent token summarizes eight raw frames and their distinct camera poses.
For raw frame $r$ and pixel $p$, let $\mathbf{T}^{\mathrm{c2w}}_r=[\mathbf{R}_r\,|\,\mathbf{o}_r]$ and $\mathbf{A}_r$ denote the raw-frame camera pose and intrinsics, with camera center $\mathbf{o}_{r}\in\mathbb{R}^{3}$ and unit ray direction $\mathbf{d}_{r,p}\in\mathbb{R}^{3}$.
We compute pixel-wise Pl\"ucker raymaps $\boldsymbol{\rho}_{r,p}=(\mathbf{d}_{r,p},\mathbf{o}_{r}\times\mathbf{d}_{r,p})\in\mathbb{R}^{6}$ from $\mathbf{T}^{\mathrm{c2w}}_r$ and $\mathbf{A}_r$.
For each latent frame, we pack the eight raw-frame raymaps within one VAE temporal stride into a $48$-channel tensor and pass it through a zero-initialized 3D patch embedder.
A zero-initialized per-block projection then adds this embedding immediately after each self-attention output, preserving the pretrained model at initialization.

\subsection{Second Stage Refiner}
\label{sec:refiner}
Following LTX-Video~\citep{ltx_video}, we add a second-stage refiner to improve stage-1 \modelname visual fidelity. 
The refiner is trained on paired latents $(\mathbf{x}_{\ell},\mathbf{x}_{h})$, where $\mathbf{x}_{\ell}$ is a stage-1 or degraded latent and $\mathbf{x}_{h}$ is the high-fidelity target. 
We use truncated-$\sigma$ flow matching: $\mathbf{x}_{\ell}$ is perturbed with a large starting noise ($\sigma_{\mathrm{start}}=0.9$), and the model learns to map this noisy source toward $\mathbf{x}_{h}$, encouraging refinement over full reconstruction. 
We condition on text, camera, and a reference image, concatenate the reference to the sequence, and exclude it from the loss to preserve stage-1 appearance. 
For minute-long refinement, we initialize from the 17B LTX-2 model and train LoRA adapters on paired synthetic data plus real videos processed by downsampling and upsampling; details are in App.~\ref{sec:appendix_refiner}.

\section{Data Pipeline}
\label{sec:data_pipeline}

\begin{figure}[t]
  \centering
  \includegraphics[width=\linewidth]{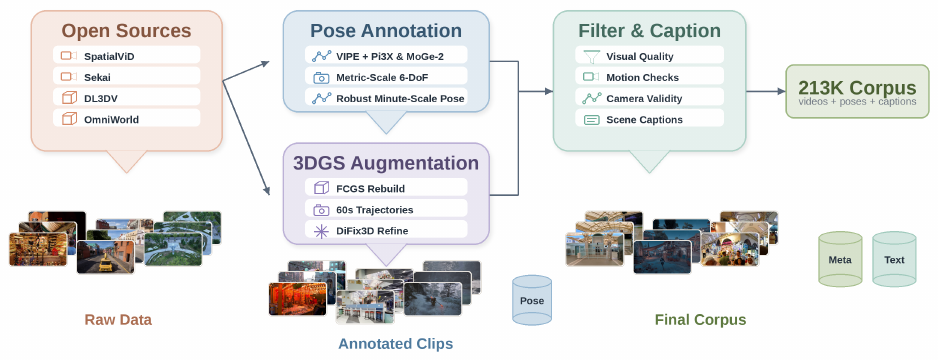}
  \caption{\textbf{Data construction pipeline.}
  We collect open-source video and static 3D sources, annotate metric-scale camera poses, augment DL3DV with 3DGS-rendered trajectories, and filter/caption the resulting clips into a 213K-clip training corpus.}
  \label{fig:data_pipeline}
\end{figure}

We build a robust annotation pipeline that re-annotates seven open-source video sources with metric-scale camera poses, yielding a 213K-clip corpus; Fig.~\ref{fig:data_pipeline} summarizes the construction flow, and Tab.~\ref{tab:data_overview} lists the resulting data.

\begin{table}[h]
  \centering
  \caption{Training data overview after filtering.}
  \label{tab:data_overview}
  \footnotesize
  \begin{tabular}{llcrc}
    \toprule
    Source & Type & Duration & Clips & Pose Source \\
    \midrule
    SpatialVID-HQ~\citep{spatialvid}  & Real           & 10\,s & 158,369 & VIPE + Pi3X/MoGe-2 \\
    DL3DV~\citep{dl3dv}               & Real           & 10\,s & 5,691   & GT pose + Pi3X \\
    DL3DV~\citep{dl3dv} GS Refined    & Synthetic      & 60s & 14,881  & GT pose + Pi3X \\
    OmniWorld~\citep{omniworld}       & Synthetic      & 60s & 1,720   & VIPE + GT depth \\
    Sekai~\citep{sekai} Game          & Synthetic      & 60s & 3,560   & GT pose + Pi3X \\
    Sekai~\citep{sekai} Walking-HQ       & Real           & 60s & 9,767   & VIPE + Pi3X/MoGe-2 \\
    MiraData~\citep{miradata}         & Real & 60s & 18,987  & VIPE + Pi3X/MoGe-2 \\
    \midrule
    \textbf{Total}       &                &       & \textbf{212,975} & \\
    \bottomrule
  \end{tabular}
\end{table}

\noindent\textbf{Sources and pose annotation.}
Our annotation engine is based on VIPE~\citep{vipe}, but we found its original depth estimation unstable on long videos.
We therefore replace the depth backend with Pi3X~\citep{pi3x} and MoGe-2~\citep{moge2}: Pi3X provides long-sequence-consistent depth, while MoGe-2 provides accurate per-frame metric scale.
This yields robust metric-scale poses for long videos in the wild. We also adapt VIPE to enable per-frame intrinsic optimization for more robust internet video annotation.
For sources with provided poses, such as Sekai-Game~\citep{sekai} and DL3DV~\citep{dl3dv}, we keep the ground-truth trajectories and use Pi3X to recover metric scale.
OmniWorld~\citep{omniworld} is a special case: because it provides ground-truth depth, we use GT depth inside VIPE and use MoGe-2 for metric-scale recovery.
Engine details are in App.~\ref{sec:appendix_vipe}.

\noindent\textbf{3DGS augmentation.}
Following HY-WorldPlay~\citep{hy_worldplay}, we augment static 3D scene data with rendered camera trajectories.
DL3DV~\citep{dl3dv} contains static 3D captures rather than native one-minute videos, so we fit one FCGS~\citep{fcgs} 3D Gaussian Splatting reconstruction per scene, design diverse one-minute camera paths, and render long videos with known intrinsics and extrinsics.
We then refine the rendered videos with DiFix3D~\citep{difix3d} to reduce splatting artifacts (App.~\ref{sec:appendix_3dgs}).

\noindent\textbf{Filtering and captioning.}
We follow SANA-Video~\citep{chen2025sana} for basic visual filtering, including aesthetic quality, motion magnitude, optical-flow consistency, and scene-cut removal.
We further add camera-specific filters on field of view, focal consistency, pose smoothness, and scale variation (App.~\ref{sec:appendix_filters}).
For captioning, we follow LingBot-World~\citep{lingbot_world}: when action conditions are available, we use scene-static captions that describe objects, layout, and appearance while omitting camera actions such as ``pan left'' or ``move forward.''
This prevents text from leaking trajectory supervision and forces motion control through the pose branch.

\section{Experiments}
\label{sec:experiments}

\subsection{Implementation Details}
\label{sec:impl_details}

\noindent\textbf{Backbone configuration.}
\modelname utilizes a Diffusion Transformer (DiT) architecture with 20 transformer blocks ($d_\text{model}{=}2240$, 20 heads). To balance global dependency modeling and efficiency, we interleave 15 frame-wise GDN blocks with softmax attention blocks at layers $\{3,7,11,15,19\}$. Every block incorporates dual UCPE + Pl\"ucker mixing camera conditioning, and the LTX2 VAE uses $C{=}128$ latent channels for high spatio-temporal compression~\citep{dit,gated_deltanet,ucpe,plucker_lfn,ltx2}.

\noindent\textbf{Training scale.}
Our final model is trained on the 213K metric-pose clips described in Sec.~\ref{sec:data_pipeline}. 
The one-minute stage is trained on 961-frame clips with the standard flow-matching objective~\citep{flow_matching}. 
For 961-frame (60s) sequences, Context-Parallel (CP) training shards the latent sequence along time. 
For efficiency, we use custom fused Triton kernels for GDN scan and gate operations. 
The model is trained on 64 H100 GPUs for approximately 15 days; architectural details, hyperparameters, and the training schedule are in App.~\ref{sec:impl_details_appendix} and App.~\ref{sec:appendix_training_hyperparams}.

\subsection{Evaluation Setup}
\label{sec:eval_setup}

\noindent
\begin{minipage}[t]{0.50\textwidth}
\vspace{0pt}
\textbf{Benchmark construction.}
For evaluation, we build a 60-second world-model benchmark with 80 initial images generated by Nano Banana Pro~\citep{google2025nanobananapro} across four scene categories: game, indoor, outdoor-city, and outdoor-nature (20 per category).
Each image is paired with Simple and Hard revisit trajectories (Fig.~\ref{fig:benchmark_trajectories}); construction details and examples are in App.~\ref{sec:appendix_benchmark_construction} and App.~\ref{sec:appendix_benchmark_examples}.

\textbf{Protocol and metrics.}
The main table uses each model's multi-step, undistilled AR setting.
We report VBench~\citep{vbench} scores for visual quality and Umeyama~\citep{umeyama}-aligned Pi3X~\citep{pi3x} pose accuracy for action following; revisit memory (PSNR/SSIM/LPIPS~\citep{lpips}) and long-term degradation $\Delta$IQ are reported in App.~\ref{sec:appendix_revisit_temporal}, with metric details in App.~\ref{sec:appendix_eval_protocol}.
\end{minipage}%
\hfill%
\begin{minipage}[t]{0.48\textwidth}
\vspace{0pt}
\centering
\includegraphics[width=\linewidth]{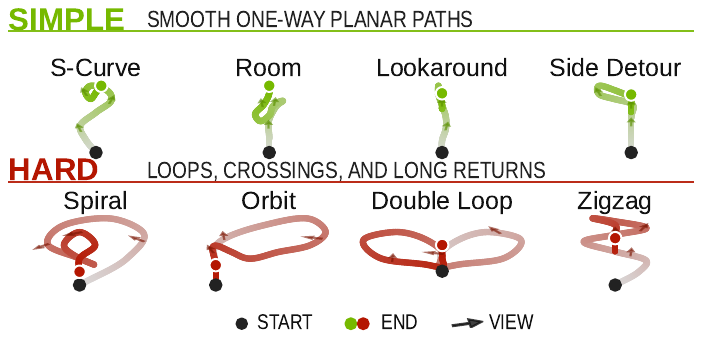}
\captionsetup{type=figure,hypcap=false}
\caption{\textbf{Representative benchmark trajectories.} BEV paths for four Simple/Hard examples.}
\label{fig:benchmark_trajectories}
\end{minipage}

\vspace{-0em}

\subsection{Main Results}
\label{sec:main_results}

\begin{table}[t]
  \centering
  \caption{Quantitative comparison on our 1-min benchmark. %
  Bold Res marks 720p. Pose Acc. reports R in degrees, plus T/CMC; VBench reports eight dimensions plus Overall. Mem/Tput are peak GB and videos/hour on 8 H100s. Green highlights mark top-three entries.}
  \label{tab:vbench}
  \resizebox{\textwidth}{!}{%
  \scriptsize
  \setlength{\tabcolsep}{1.65pt}
  \renewcommand{\arraystretch}{1.10}
  \begin{tabular}{@{} l c c c @{\hspace{4pt}} c c c @{\hspace{4pt}} c c c c c c c c c @{\hspace{4pt}} c c @{}}
    \toprule
    & & & & \multicolumn{3}{c}{\textbf{Pose Acc. ($\downarrow$)}} & \multicolumn{9}{c}{\textbf{VBench ($\uparrow$)}} & \multicolumn{2}{c}{\textbf{Efficiency}} \\
    \cmidrule(lr){5-7} \cmidrule(lr){8-16} \cmidrule(l){17-18}
    \textbf{Method} & \textbf{Param} & \textbf{Res} & \textbf{\#G}
      & \textbf{R} & \textbf{T} & \textbf{CMC}
      & \textbf{SC} & \textbf{BC} & \textbf{TF} & \textbf{MS} & \textbf{AQ} & \textbf{IQ} & \textbf{DD} & \textbf{OC} & \textbf{Overall}
      & \textbf{Mem$\downarrow$} & \textbf{Tput$\uparrow$} \\
    \midrule

    \multicolumn{18}{@{}l}{\textbf{\textit{Simple-Trajectory Split}}} \\
    \midrule
    Infinite-World~\citep{infinite_world}        & 1.3B    & 480p & 1 & 16.55 & 1.98 & 2.08 & 79.48 & 87.79 & 97.35 & 98.78 & 51.99 & 69.34 & 88.75 & 12.28 & 79.18 & \ranksecond{53.5} & \rankthird{5.9} \\
    LingBot-World~\citep{lingbot_world}          & 14B+14B & 480p & 8 & \rankthird{10.47} & 2.01 & 2.05 & 93.77 & 95.46 & 97.13 & 98.34 & 64.06 & 73.18 & 41.25 & 11.78 & \rankfirst{81.82} & 454.1 & 0.6 \\
    HY-WorldPlay~\citep{hy_worldplay}            & 8B      & 480p & 8 & 17.89 & 2.36 & 2.45 & 65.95 & 81.97 & 94.63 & 96.01 & 40.28 & 53.05 & 91.25 & 13.83 & 68.82 & 215.5 & 1.1 \\
    \midrule
    Matrix-Game\,3.0~\citep{matrixgame3}          & 5B      & \textbf{720p} & 8 & 12.96 & \rankthird{1.83} & \rankthird{1.92} & 81.62 & 90.04 & 94.37 & 97.64 & 52.24 & 66.94 & 97.50 & 13.29 & 78.53 & 106.2 & 3.1 \\
    \textbf{\modelname}               & 2.6B & \textbf{720p} & 1 & \ranksecond{7.59} & \ranksecond{1.59} & \ranksecond{1.63} & 87.46 & 91.87 & 94.99 & 97.69 & 55.70 & 69.69 & 72.50 & 11.54 & \rankthird{79.29} & \rankfirst{51.1} & \rankfirst{24.1} \\
    \textbf{\modelname + refiner}     & 2.6B+17B & \textbf{720p} & 1 & \rankfirst{4.50} & \rankfirst{1.39} & \rankfirst{1.41} & 88.62 & 93.21 & 96.18 & 98.61 & 58.05 & 72.12 & 61.25 & 11.12 & \ranksecond{80.62} & \rankthird{74.7} & \ranksecond{22.0} \\
    \midrule
    \midrule

    \multicolumn{18}{@{}l}{\textbf{\textit{Hard-Trajectory Split}}} \\
    \midrule
    Infinite-World~\citep{infinite_world}        & 1.3B    & 480p & 1 & 41.31 & 2.49 & 2.84 & 78.61 & 86.98 & 96.46 & 98.68 & 52.12 & 71.22 & 98.75 & 12.36 & 79.51 & \ranksecond{53.5} & \rankthird{5.9} \\
    LingBot-World~\citep{lingbot_world}          & 14B+14B & 480p & 8 & 18.99 & \ranksecond{1.65} & \rankthird{1.81} & 91.79 & 94.41 & 96.10 & 97.82 & 62.79 & 72.60 & 62.50 & 12.06 & \rankfirst{81.89} & 454.1 & 0.6 \\
    HY-WorldPlay~\citep{hy_worldplay}            & 8B      & 480p & 8 & 35.46 & 2.34 & 2.64 & 68.33 & 83.06 & 95.31 & 96.71 & 41.76 & 53.71 & 91.25 & 13.94 & 70.46 & 215.5 & 1.1 \\
    \midrule
    Matrix-Game\,3.0~\citep{matrixgame3}          & 5B      & \textbf{720p} & 8 & \rankthird{18.79} & 1.67 & 1.82 & 82.10 & 89.99 & 93.94 & 97.60 & 52.92 & 68.03 & 98.75 & 13.65 & 78.79 & 106.2 & 3.1 \\
    \textbf{\modelname}               & 2.6B & \textbf{720p} & 1 & \ranksecond{10.02} & \rankthird{1.66} & \ranksecond{1.72} & 85.93 & 90.89 & 94.36 & 97.49 & 53.82 & 69.12 & 92.50 & 12.10 & \rankthird{79.60} & \rankfirst{51.1} & \rankfirst{24.1} \\
    \textbf{\modelname + refiner}     & 2.6B+17B & \textbf{720p} & 1 & \rankfirst{8.34} & \rankfirst{1.39} & \rankfirst{1.44} & 87.26 & 92.55 & 95.54 & 98.49 & 56.67 & 71.38 & 91.25 & 11.34 & \ranksecond{81.89} & \rankthird{74.7} & \ranksecond{22.0} \\
    \bottomrule
  \end{tabular}}
\end{table}

\noindent\textbf{Camera control accuracy.}
Tab.~\ref{tab:vbench} compares against recent world-model baselines~\citep{infinite_world,lingbot_world,hy_worldplay,matrixgame3} and shows that \modelname gives the strongest action following on both trajectory splits.
The refined output obtains the best RotErr, TransErr, and CamMC, with RotErr $4.50^\circ$/$8.34^\circ$ and CamMC $1.41$/$1.44$, improving over both large 480p and 720p baselines.

\noindent\textbf{Visual quality.}
With the second-stage refiner, \modelname attains $80.62$/$81.89$ VBench Overall on the Simple-/Hard-Trajectory splits, close to LingBot-World ($81.82$/$81.89$) while generating 720p videos on one GPU per clip.

\noindent\textbf{Inference efficiency.}
\modelname fits in $51.1$\,GB and reaches $24.1$ videos/hour; with the refiner, the full pipeline remains within the 80\,GB H100 budget ($74.7$\,GB) and reaches $22.0$ videos/hour, still $3.7\times$ faster than the fastest visible 480p baseline.
Although LingBot-World supports 720p, minute-long 720p inference is unaffordable under our 8-H100 evaluation budget; the closest-efficiency baseline, Infinite-World, runs at 480p.

\noindent\textbf{Revisit memory and temporal stability.}
After refinement, \modelname reaches $14.46$/$14.80$\,dB revisit PSNR, ranking second on Simple and first on Hard among the visible methods.
The refiner also reduces long-horizon visual drift: $\Delta$IQ drops from $3.79$/$3.09$ for stage-1 AR outputs to $1.17$/$0.31$, avoiding the severe late-window degradation observed for HY-WorldPlay ($23.59$/$25.88$); full per-method numbers are in App.~\ref{sec:appendix_revisit_temporal}.

\begin{figure}[th]
  \centering
  \includegraphics[width=0.98\textwidth]{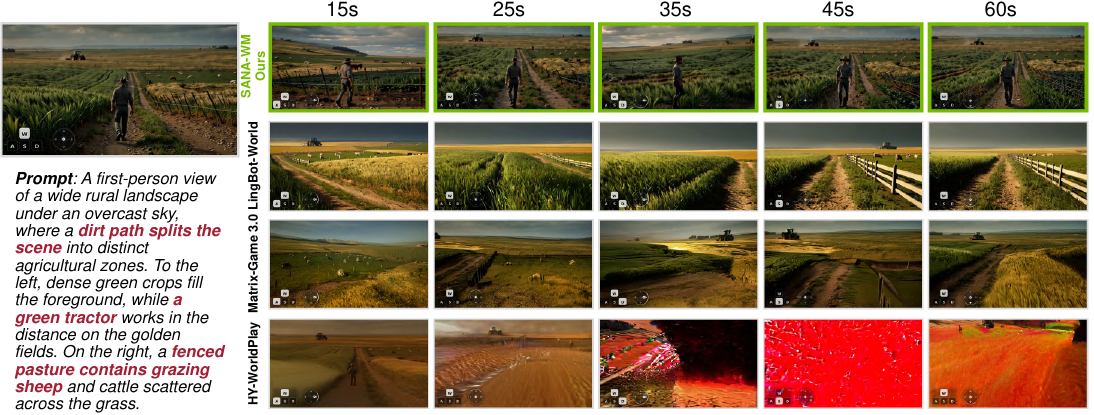}
  \vspace{3pt}
  \caption{Qualitative comparison of four Hard-Trajectory 60-second videos. Green borders denote \modelname, with transparent action overlays in the bottom-left corners. Zoom in for details.}
  \label{fig:vis-main}
\end{figure}

\noindent\modelname (Fig.~\ref{fig:vis-main}) preserves scene identity and viewpoint-consistent structure under hard actions, whereas baselines often blur, change layout, or collapse; additional examples are in App.~\ref{sec:appendix_qualitative}.

\subsection{Ablation Studies}
\label{sec:ablations}

\begin{figure}[htbp]
  \centering
  \includegraphics[width=\linewidth]{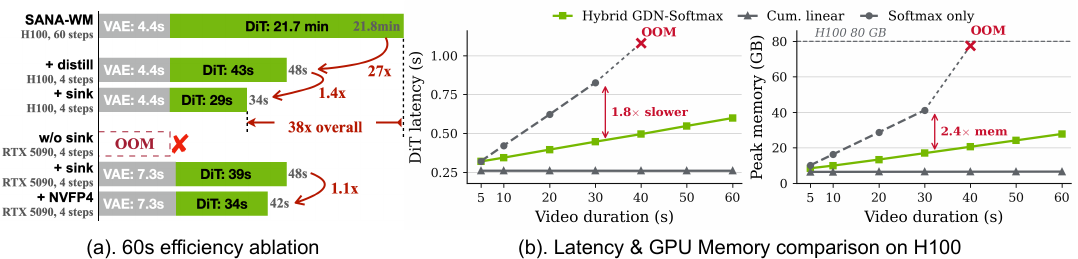}
  \vspace{2pt}
  \caption{Efficiency ablation and scaling. (a) 60s single-GPU VAE/DiT latency by stage; bars are scaled for readability. (b) H100 latency and memory scaling: recurrent variants grow compactly, while all-softmax OOMs at 60s.}
  \label{fig:efficiency-analysis}
\end{figure}

\noindent\textbf{Deployment efficiency path.} %
Fig.~\ref{fig:efficiency-analysis}~(a) isolates the internal \modelname speed path under the 60-second 720p protocol, from the 60-step H100 autoregressive generator to 4-step distillation, attention-sink deployment, and NVFP4 quantization. Note that the attention-sink variant means that we use the first latent frame as the attention sink with local window attention on softmax attention layers only, so that the memory consumption remains constant.

\noindent\textbf{Progressive-training ablation.}
Tab.~\ref{tab:ablation_stages} isolates progressive upgrades at a 5\,s horizon on VBench-I2V~\citep{vbench} with identical data and inference settings. LTX2~\citep{ltx2} is quality-neutral ($+0.0012$ Total) but cuts peak memory from $8.9$ to $5.4$\,GB and latency $3.4{\times}$; the hybrid 15-GDN/5-softmax backbone raises Total to $0.853$ while keeping memory at $5.7$\,GB.

\begin{table}[hbtp]
  \centering
  \caption{Progressive-training ablation on VBench-I2V. Scores are official rollups; efficiency is single-H100 bf16 timing, with per-dimension scores in App.~\ref{sec:appendix_ablation_stages}.}
  \label{tab:ablation_stages}
  \resizebox{\textwidth}{!}{%
  \small
  \setlength{\tabcolsep}{6pt}
  \renewcommand{\arraystretch}{1.05}
  \begin{tabular}{@{}l l l c c c c c c@{}}
    \toprule
    \textbf{Model} & \textbf{Attention} & \textbf{Tokenizer} &
      \textbf{Quality\,$\uparrow$} & \textbf{I2V\,$\uparrow$} & \textbf{Total\,$\uparrow$} &
      \textbf{Mem (GiB)\,$\downarrow$} & \textbf{Lat (ms)\,$\downarrow$} & \textbf{Tput (steps/s)\,$\uparrow$} \\
    \midrule
    Sana-Video        & cumulative linear  & Wan 2.1 / 480p & 0.7683       & 0.9073       & 0.8378       & 8.90        & 1266.6      & 0.79 \\
    + LTX2 VAE        & cumulative linear  & LTX2 / 720p    & \ranksecond{0.7697} & \ranksecond{0.9082} & \ranksecond{0.8390} & \rankfirst{5.40}  & \rankfirst{371.7} & \rankfirst{2.69} \\
    + Hybrid attn. & GDN + softmax & LTX2 / 720p    & \rankfirst{0.7834} & \rankfirst{0.9226} & \rankfirst{0.8530} & \ranksecond{5.68} & \ranksecond{433.2} & \ranksecond{2.31} \\
    \bottomrule
  \end{tabular}}
\end{table}

\input{}

\begin{figure}[htbp]
  \centering
  \begin{minipage}[t]{0.57\textwidth}
    \centering
    \vspace{2pt}
    \captionof{table}{Camera-conditioning ablation on OmniWorld. FVD~\citep{unterthiner2018towards} and Umeyama-aligned Pi3X metrics are shown.}
    \label{tab:ablation_camera}
    \vspace{0pt}
    \resizebox{0.95\linewidth}{!}{%
      \begin{tabular}{@{}lcccc@{}}
        \toprule
        Camera Encoding & FVD\,$\downarrow$ & RotErr\,$\downarrow$ & TransErr\,$\downarrow$ & CamMC\,$\downarrow$ \\
        \midrule
        No control            & 348.93 & 16.93 & 0.2347 & 0.4937 \\
        Pl\"ucker only        & 339.45 & 16.02 & 0.2340 & 0.4742 \\
        PRoPE                 & 326.70 & \ranksecond{6.29} & 0.1857 & 0.2629 \\
        UCPE only             & \rankfirst{314.88} & 7.73 & \ranksecond{0.1350} & \ranksecond{0.2453} \\
        \textbf{UCPE+Pl\"ucker} & \ranksecond{320.80} & \rankfirst{6.21} & \rankfirst{0.1162} & \rankfirst{0.2047} \\
        \bottomrule
      \end{tabular}
    }
  \end{minipage}
  \hfill
  \begin{minipage}[t]{0.40\textwidth}
    \centering
    \vspace{-5pt}
    \includegraphics[width=0.95\linewidth]{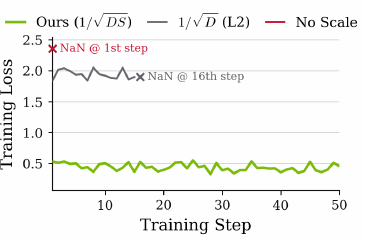}
    \vspace{-3pt}
    \captionof{figure}{Training stability ablation.}
    \label{fig:gdn-key-scaling}
  \end{minipage}
\end{figure}

\noindent\textbf{GDN key scaling.}
We evaluate training stability under identical conditions: 81-frame sequences and an all-GDN~\citep{gated_deltanet} architecture initialized from a shared LTX2-VAE~\citep{ltx2} cumulative-linear checkpoint.
As shown in Fig.~\ref{fig:gdn-key-scaling}, our $1/\sqrt{D S}$ scaling is the only variant that ensures stable convergence.
In contrast, both the $L_2$ ($1/\sqrt{D}$) and no-scale baselines suffer from immediate gradient instability, triggering NaN events at steps 16 and 1, respectively.

\noindent\textbf{Camera conditioning.}
Tab.~\ref{tab:ablation_camera} ablates camera conditioning on our held-out OmniWorld~\citep{omniworld} validation split with 5s clips after 10k fine-tuning steps. Input-level Pl\"ucker~\citep{plucker_lfn} gives only small gains, attention-level PRoPE~\citep{li2025cameras} and UCPE~\citep{ucpe} greatly improve control, and the dual UCPE + Pl\"ucker mixing branch gives the lowest Pi3X~\citep{pi3x} errors with competitive FVD~\citep{unterthiner2018towards}.

\noindent\textbf{Efficiency scaling.}
Fig.~\ref{fig:efficiency-analysis}~(b) reports AR recurrent single-step H100 latency and memory with matched resolution, camera conditioning, and FFN settings. The recurrent/linear variants keep compact clean-context states, while all-softmax grows its KV cache and runs out of memory at 60s.

\section{Conclusion}
\label{sec:conclusion}

We introduced \modelname, a camera-controlled world model for practical minute-scale 720p generation.
By combining hybrid GDN--softmax modeling, dual UCPE + Pl\"ucker conditioning, metric-pose annotation, and a long-video refiner, \modelname makes limited-compute training and single-GPU rollout compatible with 720p visual quality and 6-DoF control.
Experiments on revisit-heavy 60s trajectories show stronger camera following than open baselines and support a practical workflow: search trajectories efficiently with the stage-1 model, then refine promising rollouts for higher fidelity.

\noindent\textbf{Limitations, social impact, and future work.}
\modelname remains scale-limited, lacks explicit 3D scene memory, and can drift in dynamic scenes, rare viewpoints, or longer rollouts.
Its efficiency broadens access to simulation, embodied AI, and robotics research; practical deployment should document provenance, model scope, and evaluation settings clearly, while future work should scale models/data, explore robot action or point-tracking controls, strengthen persistent scene memory, and develop robust real-time or streaming refiners.

\newpage
\appendix
\onecolumn

\appendix

\section{Long-Video Refiner}
\label{sec:appendix_refiner}

\paragraph{Truncated-$\sigma$ flow matching.}
Given a stage-1 latent $x_l$ and its high-fidelity target $x_h$, we train the
refiner using a truncated-$\sigma$ flow-matching formulation that models the
transformation from a noised version of $x_l$ to $x_h$.

\noindent We first construct a source point by perturbing $x_l$ with Gaussian noise:
\begin{equation}
x_1 = (1 - \sigma_{\text{start}}) x_l + \sigma_{\text{start}} \epsilon,
\quad \epsilon \sim \mathcal{N}(0, I).
\end{equation}
We then sample $\sigma_t \sim p(\sigma)$ from a shifted-logit-normal
distribution truncated to $(0, \sigma_{\text{start}}]$, and define the
interpolation coefficient
\begin{equation}
    \alpha = \frac{\sigma_t}{\sigma_{\text{start}}}.
\end{equation}

\noindent The intermediate state is given by linear interpolation between the clean
target $x_h$ and the source point $x_1$:
\begin{equation}
    x_t = (1 - \alpha) x_h + \alpha x_1,
\end{equation}
where $\alpha \in (0,1]$, ensuring that all training states lie on the segment
connecting $x_h$ and $x_1$.

\noindent The target velocity is defined as
\begin{equation}
    v^\star = \frac{x_1 - x_h}{\sigma_{\text{start}}},
\end{equation}
and the model is trained to predict this velocity via a mean-squared error:
\begin{equation}
\mathcal{L}_{\text{refiner}} =
\mathbb{E}_{\sigma_t, \epsilon}
\left\| v_\theta(x_t, \sigma_t, c) - v^\star \right\|_2^2,    
\end{equation}
where $c$ denotes the conditioning inputs.

\paragraph{Reference conditioning.}
To preserve identity and appearance consistency, we prepend a clean slice of
the target latent $x_h$ (at $\sigma=0$) as reference tokens to the input
sequence. These tokens act as fixed key-value anchors via a block-wise
attention mask, and are excluded from the flow-matching loss.

\paragraph{Implementation details.}
We set $\sigma_{\text{start}} = 0.909375$, following the base model. The
noise level $\sigma_t$ is sampled from a shifted-logit-normal distribution
with truncation at $\sigma_{\text{start}}$. The refiner is trained with a
rank-384 LoRA applied to attention ($Q/K/V/O$) and feed-forward projections
under FSDP2.

\noindent The LoRA-only adaptation keeps refiner finetuning lightweight compared with
full 17B-parameter optimization. Directly finetuning the distilled few-step
refiner was unstable in our experiments, so we instead train the LoRA on the
multi-step, non-distilled LTX-2 base model and then zero-shot merge the learned
adapters into the original distilled few-step model. This transfers the
long-video behavior learned by the LoRA while preserving the distilled
inference schedule: the refiner uses the LTX-2 stage-2 distilled
sigmas $[0.909375, 0.725, 0.421875, 0]$, i.e., only three Euler denoising
steps. As a result, the refiner improves visual fidelity with limited impact
on end-to-end throughput.

\begin{figure*}[t]
    \centering
    \includegraphics[width=\textwidth]{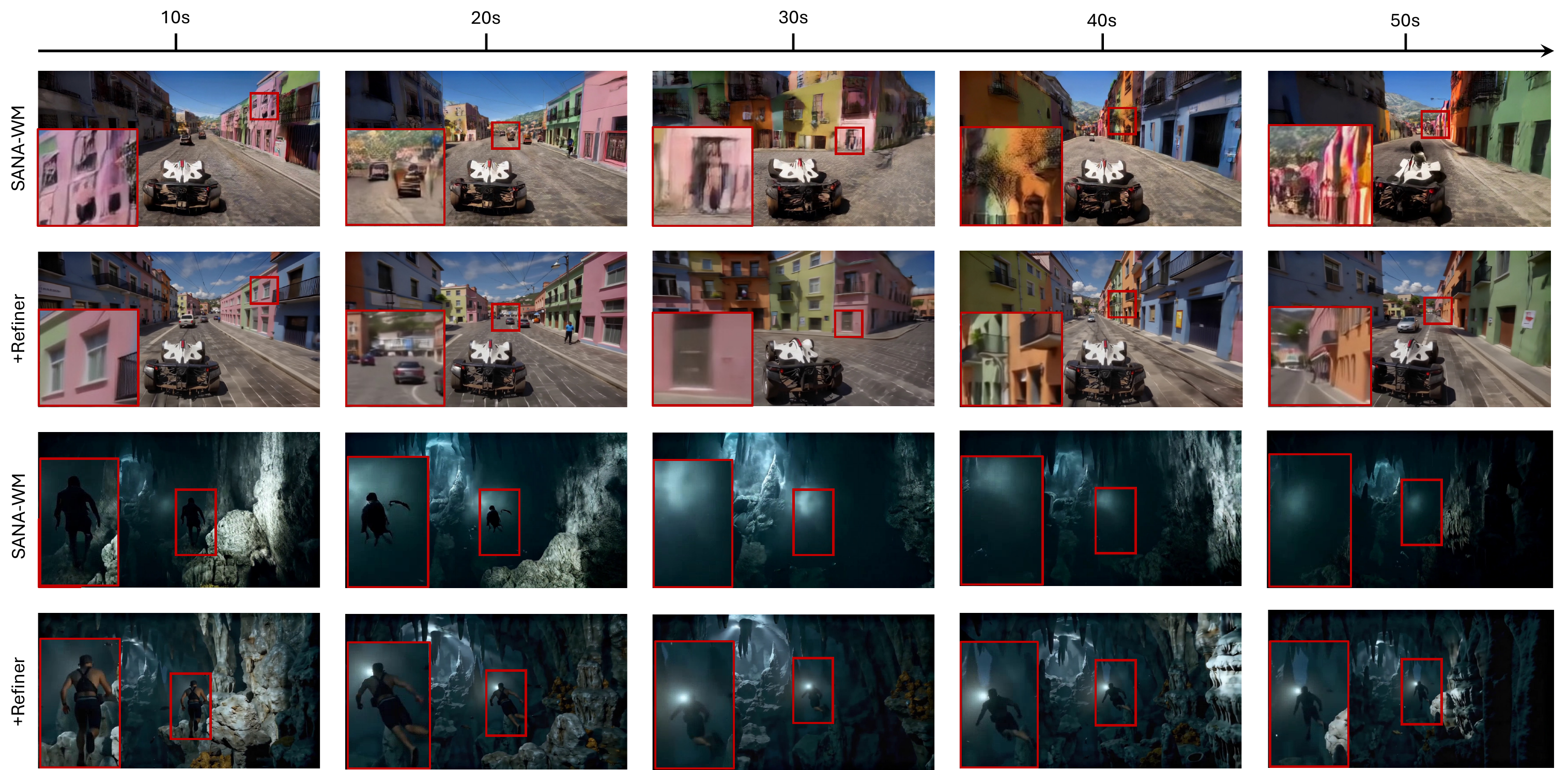}
    \caption{
    Qualitative comparison between SANA-WM and SANA-WM with the proposed refiner across long-horizon rollouts. 
    Each column shows frames sampled from 10s to 50s, and red boxes highlight local regions where the refiner improves visual fidelity, object structure, and temporal consistency.
    }
    \label{fig:appendix_vis_compare_refiner}
\end{figure*}

\noindent To isolate the effect of adapting the refiner to long \modelname videos, we run an additional ablation that reuses the same Stage-1 \modelname latents and replaces our refiner with the original LTX-2.3 refiner applied directly to the full 60-second latent sequence.
Tab.~\ref{tab:ltx23_original_refiner_ablation} shows that the original short-video refiner is not sufficient in this regime: it substantially reduces perceptual quality, late-window imaging quality, and camera-control accuracy.
The adapted refiner therefore provides the quality and control needed for minute-scale world modeling, rather than merely serving as a generic second-stage decoder.
We present the qualitative visualization results in Fig.~\ref{fig:appendix_vis_compare_refiner}.
\begin{table}[htbp]
  \centering
  \caption{Refiner ablation on the 60-second world-model benchmark using the same Stage-1 \modelname latents. VBench columns follow Tab.~\ref{tab:vbench}: SC, BC, TF, MS, AQ, IQ, DD, OC, and Overall. In Pose Acc., R is rotation error ($^\circ$), T is translation error, and CMC is camera-motion consistency. IQ$_{50\text{--}60}$ is VBench imaging quality on the last 10-second window, and $\Delta$IQ is the first-window score minus the last-window score ($\downarrow$ is more stable).}
  \label{tab:ltx23_original_refiner_ablation}
  \resizebox{\textwidth}{!}{%
  \scriptsize
  \setlength{\tabcolsep}{1.7pt}
  \renewcommand{\arraystretch}{1.10}
  \begin{tabular}{@{}l l @{\hspace{4pt}} c c c c c c c c c @{\hspace{4pt}} c c c @{\hspace{4pt}} c c@{}}
    \toprule
    & & \multicolumn{9}{c}{\textbf{VBench ($\uparrow$)}} & \multicolumn{3}{c}{\textbf{Pose Acc. ($\downarrow$)}} & \multicolumn{2}{c}{\textbf{Temporal IQ}} \\
    \cmidrule(lr){3-11} \cmidrule(lr){12-14} \cmidrule(l){15-16}
    \textbf{Split} & \textbf{Refiner}
      & \textbf{SC} & \textbf{BC} & \textbf{TF} & \textbf{MS} & \textbf{AQ} & \textbf{IQ} & \textbf{DD} & \textbf{OC} & \keycell{\textbf{Overall}}
      & \textbf{R} & \textbf{T} & \keycell{\textbf{CMC}} & IQ$_{50\text{--}60}$ & $\Delta$IQ \\
    \midrule
    \multirow{2}{*}{Simple}
      & Original LTX-2.3 & \rankfirst{94.25} & \rankfirst{95.89} & \rankfirst{99.59} & \rankfirst{99.59} & 39.73 & 38.16 & 0.00 & \rankfirst{11.39} & \keycell{71.37} & 8.65 & 2.32 & \keycell{2.35} & 35.70 & 3.73 \\
      & Ours (long-video) & 88.62 & 93.21 & 96.18 & 98.61 & \rankfirst{58.05} & \rankfirst{72.12} & \rankfirst{61.25} & 11.12 & \rankfirst{80.62} & \rankfirst{4.50} & \rankfirst{1.39} & \rankfirst{1.41} & \rankfirst{72.21} & \rankfirst{1.17} \\
    \addlinespace[2pt]
    \multirow{2}{*}{Hard}
      & Original LTX-2.3 & \rankfirst{93.15} & \rankfirst{95.78} & \rankfirst{99.60} & \rankfirst{99.60} & 40.70 & 37.17 & 0.00 & \rankfirst{11.77} & \keycell{71.16} & 27.38 & 2.29 & \keycell{2.52} & 33.69 & 4.65 \\
      & Ours (long-video) & 87.26 & 92.55 & 95.54 & 98.49 & \rankfirst{56.67} & \rankfirst{71.38} & \rankfirst{91.25} & 11.34 & \rankfirst{81.89} & \rankfirst{8.34} & \rankfirst{1.39} & \rankfirst{1.44} & \rankfirst{73.03} & \rankfirst{0.31} \\
    \bottomrule
  \end{tabular}}
\end{table}

\section{Data Pipeline Details}
\label{sec:appendix_data}

\subsection{Modified VIPE Pose Engine}
\label{sec:appendix_vipe}

We modify VIPE's~\citep{vipe} depth estimation and bundle adjustment (BA) stages while retaining its SLAM front-end (feature tracking, keyframe selection).

\noindent\textbf{Depth model upgrade.}
The original VIPE uses Metric3D-Small for single-frame depth.
We replace it with Pi3X~\citep{pi3x} (multi-frame consistent 3D structure) fused with MoGe-2~\citep{moge2} (metric-scale anchor).
The two are fused by solving for a per-frame scale factor $s$ minimizing $\sum_i w_i (s \cdot d^\text{Pi3X}_i - d^\text{MoGe}_i)^2$ with inverse-depth weights $w_i = 1/d_i$, smoothed temporally via exponential moving average (momentum $0.99$).

\noindent\textbf{Per-frame intrinsics optimization.}
The original VIPE assumes a single set of intrinsics shared across all frames.
We extend BA to treat $(f_x, f_y, c_x, c_y)$ as independent variables per frame, stored as an $(N, V, D)$ tensor (frames $\times$ views $\times$ intrinsics dimension).
Each frame's intrinsics are a separate variable in the optimization, enabling accurate calibration on internet video with non-square pixels and varying focal lengths.

\noindent\textbf{Dataset-specific annotation modes.}
(1)~\emph{Default} (internet video): full pipeline with Pi3X+MoGe-2 depth, SLAM, and per-frame BA.
(2)~\emph{GT-depth} (OmniWorld): GT depth replaces predicted depth in SLAM; MoGe-2 recovers metric scale by aligning GT point clouds.
(3)~\emph{GT-pose} (Sekai Game, DL3DV): Pi3X predicts structure; Umeyama Sim(3) alignment~\citep{umeyama} recovers the metric scale factor from GT trajectories, with 80th-percentile inlier filtering.

\subsection{3DGS Augmentation Pipeline}
\label{sec:appendix_3dgs}

Following HY-WorldPlay~\citep{hy_worldplay}, we augment static-scene datasets such as DL3DV~\citep{dl3dv} by rendering novel one-minute videos from pre-fitted FCGS~\citep{fcgs} 3D Gaussian Splats.
For each DL3DV scene, we load the FCGS reconstruction and its training cameras, compute the camera-position centroid, median scene radius, height range, and PCA directions, and use these statistics to keep synthetic cameras inside the observed capture region.
We generate 40 candidate trajectories per scene: 10 trajectories stay close to the original training views via spline interpolation over sampled camera waypoints, while 30 trajectories are drawn from diverse motion families including orbit, spiral, dolly, fly-through, random walk, crane/boom, pendulum, and compound paths.
Each trajectory is scaled by the median camera distance, anchored near a sampled training camera, reoriented toward the Gaussian centroid for forward-facing scenes, clamped to the original camera coverage zone, and smoothed with $\sigma \approx N_\text{frames}/200$.

\noindent Before rendering a full clip, we run a splat-coverage test using subsampled Gaussian centers.
Every tenth frame must project enough splats into the camera frustum for at least 70\% of sampled frames, and a $32{\times}32$ tile coverage check rejects views where more than 65\% of image tiles are empty.
After FCGS rendering, we additionally discard clips with more than 30\% near-blank frames.
Finally, rendered clips are refined frame-by-frame with the no-reference DiFix3D~\citep{difix3d} pipeline (single diffusion step, prompt ``remove degradation'', timestep 199, guidance scale 0).
The refined set is further filtered based on saturation, scene cuts, VMAF motion, and first-frame black-pixel ratio.

\subsection{Per-Dataset Quality Filter Thresholds}
\label{sec:appendix_filters}

We apply a unified scoring pipeline before training-data selection.
For visual and motion quality, we measure mean color saturation, FFmpeg VMAF motion, UniMatch optical-flow magnitude, DOVER technical/aesthetic quality, and scene cuts from PySceneDetect.
For long videos, UniMatch samples frame pairs every 0.5s across the first 60s, and DOVER is averaged over non-overlapping 5s chunks.
We also run a Qwen3.5 VLM~\citep{qwen3_5} pass to count people, vehicles, and animals in each scene and to flag clips that are visually poor, overly dark, or otherwise unsuitable.
Tab.~\ref{tab:filter_thresholds} summarizes the per-dataset thresholds; each metric value must fall within the specified range for a clip to be retained.
\textbf{VLM Entity} denotes the allowed count of people, vehicles, and animals, while \textbf{VLM Quality} denotes the accepted range of the VLM quality flag.
In addition, camera-specific filters are applied uniformly across all datasets.
Given frame resolution $(W,H)$ and intrinsics $(f_x,f_y,c_x,c_y)$, we compute horizontal and vertical fields of view as $\theta_x=2\arctan(W/(2f_x))$ and $\theta_y=2\arctan(H/(2f_y))$, and require both to lie in $[25^\circ,120^\circ]$.
The $f_x/f_y$ divergence is the symmetric normalized focal mismatch,
$|f_x-f_y| / \bigl((f_x+f_y)/2\bigr)$, which must be at most $0.20$.
For metric scale, let $\{s_t\}_{t=1}^T$ be the per-frame scale factors estimated during pose annotation; we compute the coefficient of variation as $\mathrm{std}(s_t)/(\mathrm{mean}(s_t)+\epsilon)$ and reject clips with value above $2.0$.

\begin{table}[htbp]
\centering
\caption{Per-dataset quality filter thresholds. Ranges denote $[\min, \max]$ acceptable values. ``---'' indicates the filter is not applied for that dataset.}
\label{tab:filter_thresholds}
\scriptsize
\begin{tabular}{@{}lccccccc@{}}
\toprule
\textbf{Dataset} & \textbf{VMAF Motion} & \textbf{UniMatch} & \textbf{DOVER} & \textbf{Color Sat.} & \textbf{Scene Cuts} & \textbf{VLM Entity} & \textbf{VLM Quality} \\
\midrule
OmniWorld       & $[0.5, 100]$ & $[3, 100]$ & $[0.35, 1.0]$ & ---          & ---    & $[0, 10]$  & $[0.5, 1.5]$ \\
Sekai Game      & $[0.5, 50]$  & $[3, 80]$  & $[0.25, 1.0]$ & ---          & ---    & $[0, 10]$  & $[0.5, 1.5]$ \\
Sekai Walking   & $[0.5, 50]$  & $[3, 50]$  & $[0.35, 1.0]$ & $[0, 180]$   & ---    & $[0, 25]$  & $[0.5, 1.5]$ \\
MiraData        & $[0.5, 50]$  & $[3, 80]$  & $[0.4, 1.0]$  & $[0, 180]$   & $\leq 1$ & ---        & --- \\
DL3DV-GS        & $[6, 50]$    & $[3, 80]$  & $[0.4, 1.0]$  & $[0, 180]$   & $\leq 1$ & ---        & --- \\
SpatialVID      & $[0.5, 50]$  & $[3, 80]$  & $[0.35, 1.0]$ & $[0, 180]$   & ---    & $[0, 10]$  & $[0.5, 1.5]$ \\
\bottomrule
\end{tabular}
\end{table}

\section{Implementation Details}
\label{sec:appendix}

\subsection{Detailed Implementation and Training Efficiency}
\label{sec:impl_details_appendix}

\noindent \textbf{Architectural Specifics.}
The \modelname backbone is designed for long-horizon stability. The 20 transformer blocks use a head dimension of $D{=}112$ and interleave 15 frame-wise GDN blocks with softmax blocks at $\{3,7,11,15,19\}$. Every block uses dual UCPE + Pl\"ucker mixing camera conditioning. The interleaved softmax blocks allow the model to anchor long-term spatial consistency, while the GDN blocks provide efficient frame-by-frame evolution.

\noindent \textbf{Context-Parallel (CP) Training}.
To make 961-frame training feasible, we shard the latent sequence across $P$ GPUs. Each rank $p$ holds frames $\mathcal{I}_p=\{pT/P,\ldots,(p{+}1)T/P{-}1\}$. Since the GDN update (Eq.~\ref{eq:frame_gdn}) is affine, each shard computes a transition composite $\mathbf{C}_p$ and an input composite $\mathbf{H}_p$:

\begin{equation}
\mathbf{S}_{\mathrm{end}}^{(p)}=\mathbf{S}_{\mathrm{start}}^{(p)}\mathbf{C}_p+\mathbf{H}_p,
\quad
\text{where}
\quad
\mathbf{C}_p=\prod_{t\in\mathcal{I}_p}\mathbf{M}_t .
\end{equation}

\noindent We all-gather these compact summaries (rather than full activations) and compose them as an exclusive prefix:

\begin{equation}
\bar{\mathbf{S}}_0=\mathbf{0},
\quad
\bar{\mathbf{S}}_{p+1}=\bar{\mathbf{S}}_p\mathbf{C}_p+\mathbf{H}_p,
\quad
\mathbf{S}_{\mathrm{start}}^{(p)}=\bar{\mathbf{S}}_p.
\end{equation}

\noindent This recovers the mathematically exact initial GDN state for each rank with minimal communication overhead.

\noindent \textbf{Halo Exchange for Convolutions.}
While the recurrent scan uses prefix-sum composition, the temporal convolutions in GDN and FFN blocks require boundary context. For a kernel size $K$, rank $p$ exchanges $K-1$ frames with its neighbors to form a halo-augmented sequence $\tilde{\mathbf{X}}^{(p)}$, ensuring the output $\mathbf{Y}^{(p)}$ matches the unsharded version exactly. For chunk causal finetuning, we use zero padding at the global sequence boundaries and enforce causality by omitting the right-side halo where applicable.

\noindent \textbf{Fused Triton Kernels.} We implement high-performance kernels using OpenAI Triton to fuse multiple operations: RMSNorm, ReLU, key scaling, UCPE/RoPE preparation, and the GDN recurrent scan. This fusion significantly reduces memory bandwidth bottlenecks and provides autograd support, contributing to about $1.5\times$ to $2\times$ efficiency gain.

\subsection{Training Hyperparameters}
\label{sec:appendix_training_hyperparams}

Before DiT training, we adapt the LTX2 VAE on the SANA-Video SFT training data for about 50K steps, which takes roughly 3.5 days on 64 H100 GPUs. The main DiT training stages and compute budgets are summarized in Tab.~\ref{tab:training_stages}; together, they require about 15 days on 64 H100 GPUs. During minute-scale long-video training (Stages 3 and 4), we precompute VAE latents to remove the cost of online VAE encoding. Stages 1--2 use a per-GPU batch size of 1, while CP size 2 in Stages 3--4 corresponds to 0.5 clips per GPU and an effective global batch size of 32 on 64 GPUs.

\begin{table}[htbp]
  \centering
  \caption{Training schedule and hyperparameters for the progressive DiT/backbone training pipeline. All stages use AdamW, BF16 mixed precision, and gradient clipping at 0.5.}
  \label{tab:training_stages}
  \resizebox{\textwidth}{!}{%
  \begin{tabular}{lcccc}
    \toprule
    & \textbf{Stage 1} & \textbf{Stage 2} & \textbf{Stage 3} & \textbf{Stage 4} \\
    \midrule
    Purpose          & Frame-wise GDN      & Hybrid Attention   & Minute-Scale Video + CamCtrl & SFT \\
    Training data    & SANA-Video SFT data & SANA-Video SFT data & \modelname data (Sec.~\ref{sec:data_pipeline}) & $\sim$50K high-quality clips \\
    Clip duration    & 5s                  & 5s                  & 1 min               & 1 min \\
    Batch/GPU        & 1                   & 1                   & 0.5                 & 0.5 \\
    CP size          & --                  & --                  & 2                   & 2 \\
    Effective global batch & 64             & 64                  & 32                  & 32 \\
    Learning rate    & $5 \times 10^{-5}$  & $5 \times 10^{-5}$  & $1 \times 10^{-5}$  & $1 \times 10^{-5}$ \\
    Training steps   & 30K                 & 30K                 & 31K                 & 10K \\
    Compute budget   & $\sim$2.75 days     & $\sim$2 days        & $\sim$8 days        & $\sim$2.5 days \\
    \bottomrule
  \end{tabular}
  }
\end{table}

\section{Benchmark Details}
\label{sec:appendix_benchmark}

\subsection{Benchmark Construction}
\label{sec:appendix_benchmark_construction}

Our benchmark starts from 80 first-frame conditioning images at $1280{\times}720$ resolution, generated by Nano Banana Pro~\citep{google2025nanobananapro} and balanced across four scene categories: game-style, indoor, outdoor-city, and outdoor-nature (20 scenes each).
Each scene is then annotated by a Qwen3.5 VLM~\citep{qwen3_5} with a scene-static first-person prompt that describes layout, materials, lighting, and autonomous world dynamics while avoiding camera-motion language.

\noindent We use two trajectory splits.
The Simple split contains 10 smooth navigation templates with arcs, S-curves, backtracking, and figure-eight returns.
The Hard split contains 10 stress-test templates with larger yaw changes, vertical motion, extreme pitch, whip-pan events, double loops, and crane-like moves.
Each template is specified as 11--15 waypoints with position, yaw, pitch, time, and revisit metadata.
Given a target duration and frame count, we interpolate positions with centripetal Catmull-Rom splines and rotations with quaternion Squad interpolation.
We then apply arc-length reparameterization for approximately uniform speed, Laplacian position smoothing, and angular-velocity clamping at $12^\circ$/s, producing smooth frame-level camera trajectories rather than piecewise constant action chunks.
The main benchmark uses the 60-second versions of both splits: each split contains 80 scenes at 16\,fps, giving 960 video frames and 961 \modelname camera frames.
For each generated path, we also store revisit pairs where the camera nearly returns to the same viewpoint (position distance $<0.5$\,m and viewing-angle difference $<20^\circ$); these pairs support the revisit-memory metric described below.

\noindent To adapt these template trajectories to each scene and avoid invalid camera motion, we estimate camera intrinsics and metric depth from the conditioning image with Pi3X~\citep{pi3x}.
The resulting median depth provides a scene-scale proxy: travel distance is limited to 60\% of the median depth, with a global maximum speed of $0.4$\,m/s, so indoor scenes move more slowly while outdoor scenes can traverse farther.
Pi3X also provides a downsampled metric point cloud for boundary checking.
For collision avoidance, each candidate trajectory is checked against this point cloud with a 0.3\,m margin; if a camera center collides with the scene, the maximum speed is reduced by 30\% and the trajectory is regenerated for up to three retries.
For reproducibility, the benchmark package records scene metadata, prompts, calibrated intrinsics, metric camera trajectories, collision status, smoothness statistics, revisit pairs, and model-specific action or camera-control inputs generated from the same underlying trajectories.

\subsection{Benchmark Examples}
\label{sec:appendix_benchmark_examples}

\begin{figure}[htbp]
  \centering
  \includegraphics[width=\linewidth]{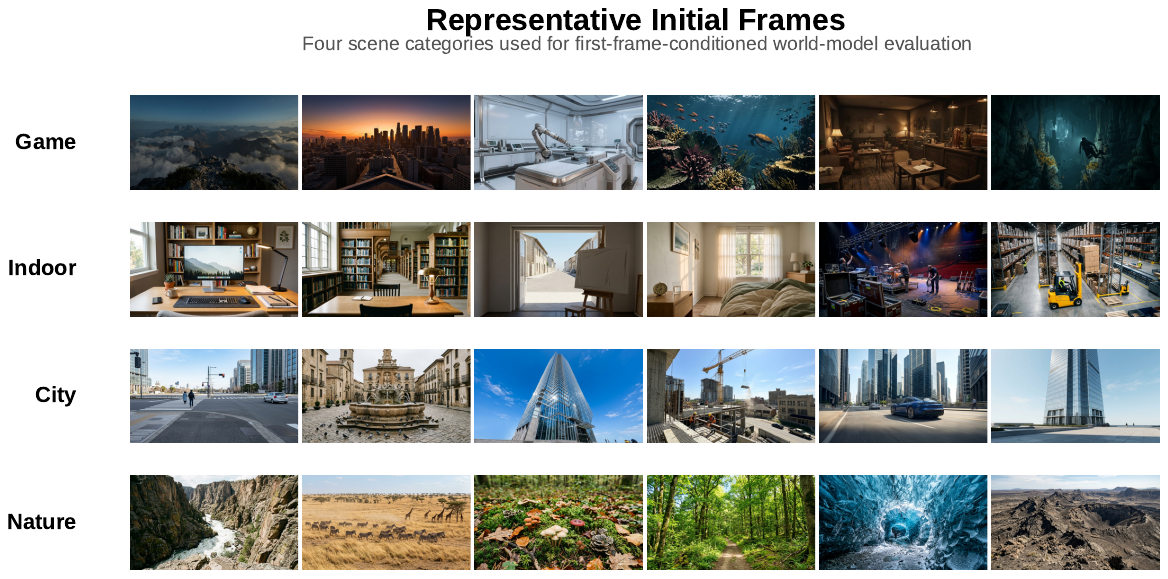}
  \caption{Representative first-frame conditioning images from the benchmark. Each row shows examples from one scene category, illustrating the diversity of geometry, lighting, and visual style used for evaluation; zooming in reveals the fine-grained scene details.}
  \label{fig:appendix_benchmark_initial_frames}
\end{figure}

\begin{figure}[htbp]
  \centering
  \includegraphics[width=\linewidth]{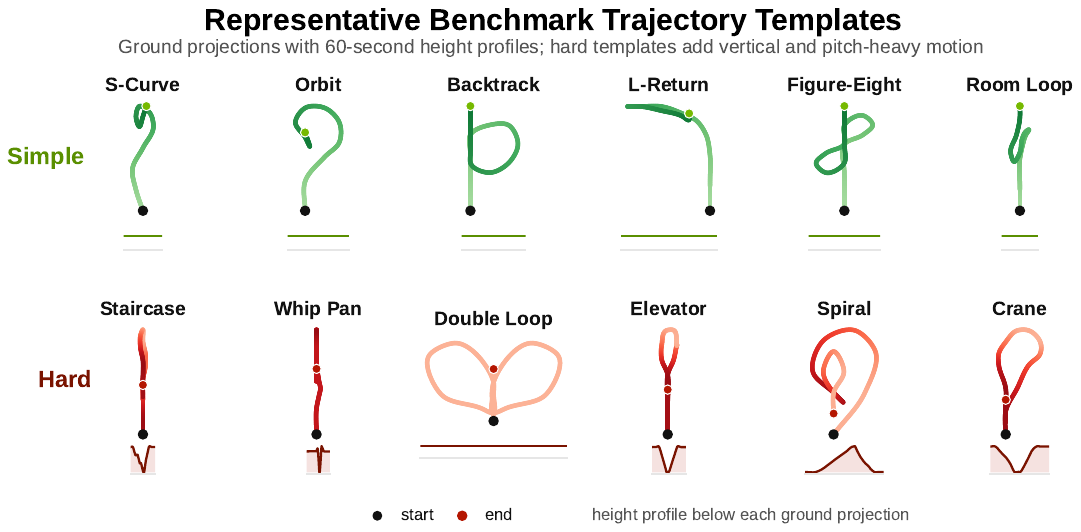}
  \caption{Representative benchmark trajectory templates. The Simple split includes smooth symmetric or revisiting planar paths, while the Hard split adds vertical motion, loop closures, and pitch-heavy viewpoints. Each panel shows the ground-plane projection with a 60-second camera-height profile below, making out-of-plane motion visible beyond a BEV path alone.}
  \label{fig:appendix_benchmark_hard_3d}
\end{figure}

\subsection{Evaluation Protocol}
\label{sec:appendix_eval_protocol}

All methods are evaluated from one generated video per benchmark scene under the same split-specific protocol.
Because different baselines use different native frame rates, revisit evaluation remaps frame indices by timestamp: a reference frame index $i$ at 16\,fps is read from generated frame $\mathrm{round}(i\,f_{\mathrm{video}}/16)$.
The benchmark exports model-specific camera-control inputs from the same underlying trajectories, including discrete action labels, pose tracks, camera-control tensors, and calibrated intrinsics.
This keeps first frames, prompts, and trajectories fixed while allowing each baseline to run in its native interface.

\noindent\textbf{Visual quality.}
We run VBench~\citep{vbench} in custom-input mode using the benchmark prompts.
For the main tables, we report the video-quality and prompt-consistency dimensions that are valid for custom-input evaluation: subject consistency (SC), background consistency (BC), temporal flickering (TF), motion smoothness (MS), aesthetic quality (AQ), imaging quality (IQ), dynamic degree (DD), and overall consistency (OC), together with the aggregated VBench Overall score.
Scores are cached per dimension and aggregated following the VBench normalization and weighting convention, with dynamic degree weighted by $0.5$ in the quality group.

\noindent\textbf{Camera accuracy.}
We estimate camera poses from each generated video using Pi3X~\citep{pi3x}.
The ground-truth trajectory is loaded from the benchmark camera annotations, relativized to the first frame, and sampled consistently with the evaluated video.
The recovered Pi3X trajectory is aligned to the ground truth by Umeyama Sim(3) alignment~\citep{umeyama}.
Let $\mathbf{P}_t=[\mathbf{R}_t|\mathbf{t}_t]$ be the ground-truth pose and $\widetilde{\mathbf{P}}_t=[\widetilde{\mathbf{R}}_t|\widetilde{\mathbf{t}}_t]$ be the aligned Pi3X pose for frame $t$.
We report
\[
\mathrm{RotErr}=\frac{1}{T}\sum_{t=1}^{T}\frac{180}{\pi}
\cos^{-1}\!\left(\operatorname{clip}\!\left(\frac{\operatorname{tr}(\mathbf{R}_t^\top\widetilde{\mathbf{R}}_t)-1}{2},-1,1\right)\right),
\]
\[
\mathrm{TransErr}=\frac{1}{T}\sum_{t=1}^{T}\|\mathbf{t}_t-\widetilde{\mathbf{t}}_t\|_2,\qquad
\mathrm{CamMC}=\frac{1}{T}\sum_{t=1}^{T}\|\mathbf{P}_t-\widetilde{\mathbf{P}}_t\|_F,
\]
where $\|\cdot\|_F$ is the Frobenius norm, RotErr is in degrees, and all three metrics are lower-is-better.

\noindent\textbf{Revisit memory.}
For each scene, we sort the stored revisit pairs by their quality score and evaluate at most five closest pairs.
Each pair compares two generated frames at nearly identical camera viewpoints using PSNR, SSIM, and LPIPS~\citep{lpips}; higher PSNR/SSIM and lower LPIPS indicate better scene memory under loop closure.
The reported split-level numbers are averages over all valid pairs and scenes.

\noindent\textbf{Temporal degradation.}
To measure long-horizon visual drift, we split each 60-second video into non-overlapping 10-second windows and run VBench on each window.
The main temporal-stability statistic is $\Delta$IQ, the imaging-quality score in the first window minus the imaging-quality score in the last window; lower values indicate less degradation over the minute.

\section{Additional Results}
\label{sec:appendix_results}

\subsection{Revisit memory and temporal stability (full table)}
\label{sec:appendix_revisit_temporal}

Tab.~\ref{tab:revisit_temporal} expands the brief revisit and temporal-stability discussion in Sec.~\ref{sec:main_results}.
Revisit metrics test whether a model preserves scene identity when the camera returns to a nearby viewpoint, while $\Delta$IQ measures late-window visual degradation over a full minute.
The results show that \modelname remains competitive on loop-closure memory at 720p, and the second-stage refiner consistently reduces long-horizon drift, especially on the Hard-Trajectory split.

\begin{table}[t]
  \centering
  \caption{Revisit memory and temporal stability on the 60-second benchmark. PSNR/SSIM/LPIPS use same-pose pairs; $\Delta$IQ compares VBench imaging quality between the first and last 10-second windows. Bold Res marks 720p, and green highlights mark top-three entries per split.}
  \label{tab:revisit_temporal}
  \scriptsize
  \setlength{\tabcolsep}{2.5pt}
  \renewcommand{\arraystretch}{1.10}
  \begin{tabular}{@{} l c c c @{\hspace{5pt}} c c c @{\hspace{5pt}} c c c @{}}
    \toprule
    & & & & \multicolumn{3}{c}{\textbf{Revisit consistency}} & \multicolumn{3}{c}{\textbf{Temporal degradation}} \\
    \cmidrule(lr){5-7} \cmidrule(l){8-10}
    \textbf{Method} & \textbf{Param} & \textbf{Res} & \textbf{\#G} & \textbf{PSNR$\uparrow$} & \textbf{SSIM$\uparrow$} & \textbf{LPIPS$\downarrow$} &
                      \textbf{IQ$_{0\text{--}10}\uparrow$} & \textbf{IQ$_{50\text{--}60}\uparrow$} & \textbf{$\Delta$IQ$\downarrow$} \\
    \midrule
    \multicolumn{10}{@{}l}{\textbf{\textit{Simple-Trajectory Split}}} \\
    \midrule
    Infinite-World~\citep{infinite_world}       & 1.3B    & 480p & 1 & 12.60 & 0.284 & 0.595 & \rankfirst{73.93} & 67.22 & \phantom{+}6.72 \\
    LingBot-World~\citep{lingbot_world}         & 14B+14B & 480p & 8 & \rankfirst{14.59} & \rankfirst{0.366} & \rankfirst{0.394} & \ranksecond{73.46} & \rankfirst{73.42} & \rankfirst{\phantom{+}0.04} \\
    HY-WorldPlay~\citep{hy_worldplay}           & 8B      & 480p & 8 & 12.83 & 0.321 & 0.616 & 70.08 & 46.50 & \phantom{+}23.59 \\
    \midrule
    Matrix-Game\,3.0~\citep{matrixgame3}        & 5B      & \textbf{720p} & 8 & 12.29 & \rankthird{0.326} & 0.553 & 69.07 & 66.66 & \rankthird{\phantom{+}2.41} \\
    \textbf{\modelname}                         & 2.6B    & \textbf{720p} & 1 & \rankthird{14.16} & \ranksecond{0.333} & \ranksecond{0.458} & 72.63 & \rankthird{68.84} & \phantom{+}3.79 \\
    \textbf{\modelname + refiner}               & 2.6B    & \textbf{720p} & 1 & \ranksecond{14.46} & 0.292 & \rankthird{0.479} & \rankthird{73.37} & \ranksecond{72.21} & \ranksecond{\phantom{+}1.17} \\
    \midrule
    \midrule
    \multicolumn{10}{@{}l}{\textbf{\textit{Hard-Trajectory Split}}} \\
    \midrule
    Infinite-World~\citep{infinite_world}       & 1.3B    & 480p & 1 & 12.04 & 0.248 & 0.617 & \rankfirst{73.79} & \rankthird{69.63} & \phantom{+}4.16 \\
    LingBot-World~\citep{lingbot_world}         & 14B+14B & 480p & 8 & \rankthird{14.08} & \rankfirst{0.332} & \rankfirst{0.436} & \ranksecond{73.66} & \rankfirst{73.09} & \rankthird{\phantom{+}0.58} \\
    HY-WorldPlay~\citep{hy_worldplay}           & 8B      & 480p & 8 & 13.72 & \ranksecond{0.328} & 0.654 & 70.21 & 44.33 & \phantom{+}25.88 \\
    \midrule
    Matrix-Game\,3.0~\citep{matrixgame3}        & 5B      & \textbf{720p} & 8 & 12.17 & 0.317 & 0.556 & 69.24 & 68.92 & \ranksecond{\phantom{+}0.32} \\
    \textbf{\modelname}                         & 2.6B    & \textbf{720p} & 1 & \ranksecond{14.10} & \rankthird{0.327} & \rankthird{0.469} & 72.58 & 69.49 & \phantom{+}3.09 \\
    \textbf{\modelname + refiner}               & 2.6B    & \textbf{720p} & 1 & \rankfirst{14.80} & 0.312 & \ranksecond{0.458} & \rankthird{73.34} & \ranksecond{73.03} & \rankfirst{\phantom{+}0.31} \\
    \bottomrule
  \end{tabular}
\end{table}

\begin{table}[t]
  \centering
  \caption{\modelname bidirectional and autoregressive stage-1 comparison on the 60-second benchmark. The top panel follows the main quantitative table; the bottom panel reports the revisit-memory and temporal-stability metrics used in Tab.~\ref{tab:revisit_temporal}. Both variants use the same 2.6B 720p backbone and are measured as single-GPU rollouts for efficiency.}
  \label{tab:sanawm_bidir_ar_full}
  \scriptsize
  \setlength{\tabcolsep}{1.5pt}
  \renewcommand{\arraystretch}{1.10}
  \begin{tabular}{@{} l c c c c @{\hspace{4pt}} c c c @{\hspace{4pt}} c c c c c c c c c @{\hspace{4pt}} c c @{}}
    \toprule
    & & & & & \multicolumn{3}{c}{\textbf{Pose Acc. ($\downarrow$)}} & \multicolumn{9}{c}{\textbf{VBench ($\uparrow$)}} & \multicolumn{2}{c}{\textbf{Efficiency}} \\
    \cmidrule(lr){6-8} \cmidrule(lr){9-17} \cmidrule(l){18-19}
    \textbf{Split} & \textbf{Mode} & \textbf{Param} & \textbf{Res} & \textbf{\#G}
      & \textbf{R} & \textbf{T} & \textbf{CMC}
      & \textbf{SC} & \textbf{BC} & \textbf{TF} & \textbf{MS} & \textbf{AQ} & \textbf{IQ} & \textbf{DD} & \textbf{OC} & \textbf{Overall}
      & \textbf{Mem$\downarrow$} & \textbf{Tput$\uparrow$} \\
    \midrule
    Simple & Bidir. & 2.6B & \textbf{720p} & 1 & 3.11 & 1.08 & 1.09 & 90.34 & 93.31 & 95.65 & 98.24 & 58.54 & 72.32 & 45.00 & 11.63 & 79.39 & 49.2 & 29.5 \\
    Simple & AR     & 2.6B & \textbf{720p} & 1 & 7.59 & 1.59 & 1.63 & 87.46 & 91.87 & 94.99 & 97.69 & 55.70 & 69.69 & 72.50 & 11.54 & 79.29 & 51.1 & 24.1 \\
    Hard   & Bidir. & 2.6B & \textbf{720p} & 1 & 3.17 & 1.08 & 1.09 & 87.23 & 91.50 & 94.92 & 97.96 & 55.02 & 70.56 & 83.75 & 11.93 & 80.18 & 49.2 & 29.5 \\
    Hard   & AR     & 2.6B & \textbf{720p} & 1 & 10.02 & 1.66 & 1.72 & 85.93 & 90.89 & 94.36 & 97.49 & 53.82 & 69.12 & 92.50 & 12.10 & 79.60 & 51.1 & 24.1 \\
    \bottomrule
  \end{tabular}

  \vspace{2pt}
  \scriptsize
  \setlength{\tabcolsep}{4pt}
  \renewcommand{\arraystretch}{1.08}
  \begin{tabular}{@{} l c @{\hspace{6pt}} c c c @{\hspace{6pt}} c c c @{}}
    \toprule
    & & \multicolumn{3}{c}{\textbf{Revisit consistency}} & \multicolumn{3}{c}{\textbf{Temporal degradation}} \\
    \cmidrule(lr){3-5} \cmidrule(l){6-8}
    \textbf{Split} & \textbf{Mode} & \textbf{PSNR$\uparrow$} & \textbf{SSIM$\uparrow$} & \textbf{LPIPS$\downarrow$} &
    \textbf{IQ$_{0\text{--}10}\uparrow$} & \textbf{IQ$_{50\text{--}60}\uparrow$} & \textbf{$\Delta$IQ$\downarrow$} \\
    \midrule
    Simple & Bidir. & 13.74 & 0.346 & 0.425 & 73.50 & 71.25 & \phantom{+}2.25 \\
    Simple & AR     & 14.16 & 0.333 & 0.458 & 72.63 & 68.84 & \phantom{+}3.79 \\
    Hard   & Bidir. & 13.78 & 0.342 & 0.432 & 72.95 & 70.82 & \phantom{+}2.13 \\
    Hard   & AR     & 14.10 & 0.327 & 0.469 & 72.58 & 69.49 & \phantom{+}3.09 \\
    \bottomrule
  \end{tabular}
\end{table}

\subsection{Progressive-training ablation: VBench-I2V per-dimension breakdown}
\label{sec:appendix_ablation_stages}

Tab.~\ref{tab:ablation_stages_perdim} expands the rolled-up scores of Tab.~\ref{tab:ablation_stages} into the nine VBench-I2V dimensions, split into the three I2V-specific axes (\emph{camera\_motion}, \emph{i2v\_subject}, \emph{i2v\_background}) and the six general video-quality axes (\emph{subject\_consistency}, \emph{background\_consistency}, \emph{motion\_smoothness}, \emph{dynamic\_degree}, \emph{aesthetic\_quality}, \emph{imaging\_quality}). The decomposition shows that the LTX2 VAE swap preserves quality while greatly improving efficiency, and the hybrid backbone improves the long-range consistency and I2V dimensions beyond the starting Sana-Video baseline.

\begin{table}[htbp]
  \centering
  \caption{VBench-I2V per-dimension scores ($\uparrow$, $0$--$1$) for the visible progressive-training stages of Tab.~\ref{tab:ablation_stages}. Same checkpoints, prompts, seeds, sampler, and inference settings as the main table. \textbf{I2V-specific dims}: \emph{camera\_motion} (CM), \emph{i2v\_subject} (IS), \emph{i2v\_background} (IB). \textbf{General quality dims}: \emph{subject\_consistency} (SC), \emph{background\_consistency} (BC), \emph{motion\_smoothness} (MS), \emph{dynamic\_degree} (DD), \emph{aesthetic\_quality} (AQ), \emph{imaging\_quality} (IQ). Green highlights mark the best value per column.}
  \label{tab:ablation_stages_perdim}
  \scriptsize
  \setlength{\tabcolsep}{4pt}
  \renewcommand{\arraystretch}{1.05}
  \begin{tabular}{@{}l ccc ccc ccc@{}}
    \toprule
    & \multicolumn{3}{c}{\textbf{I2V-specific}} & \multicolumn{6}{c}{\textbf{General quality}} \\
    \cmidrule(lr){2-4} \cmidrule(l){5-10}
    \textbf{Model} & CM & IS & IB & SC & BC & MS & DD & AQ & IQ \\
    \midrule
    Sana-Video        & \rankfirst{0.4755} & 0.9312       & 0.9545       & 0.8267       & 0.8841       & \rankfirst{0.9613} & \rankfirst{0.9976} & \rankfirst{0.5738} & 0.6376 \\
    + LTX2 VAE        & 0.3942       & 0.9309       & 0.9622       & 0.8346       & 0.9048       & 0.9560       & 0.9902       & 0.5395       & 0.6636 \\
    + Hybrid attention& 0.4343       & \rankfirst{0.9450} & \rankfirst{0.9693} & \rankfirst{0.8564} & \rankfirst{0.9114} & 0.9611       & 0.9602       & 0.5649       & \rankfirst{0.6765} \\
    \bottomrule
  \end{tabular}
\end{table}

\subsection{Additional Qualitative Results}
\label{sec:appendix_qualitative}

\begin{figure}[t]
  \centering
  \includegraphics[width=\textwidth]{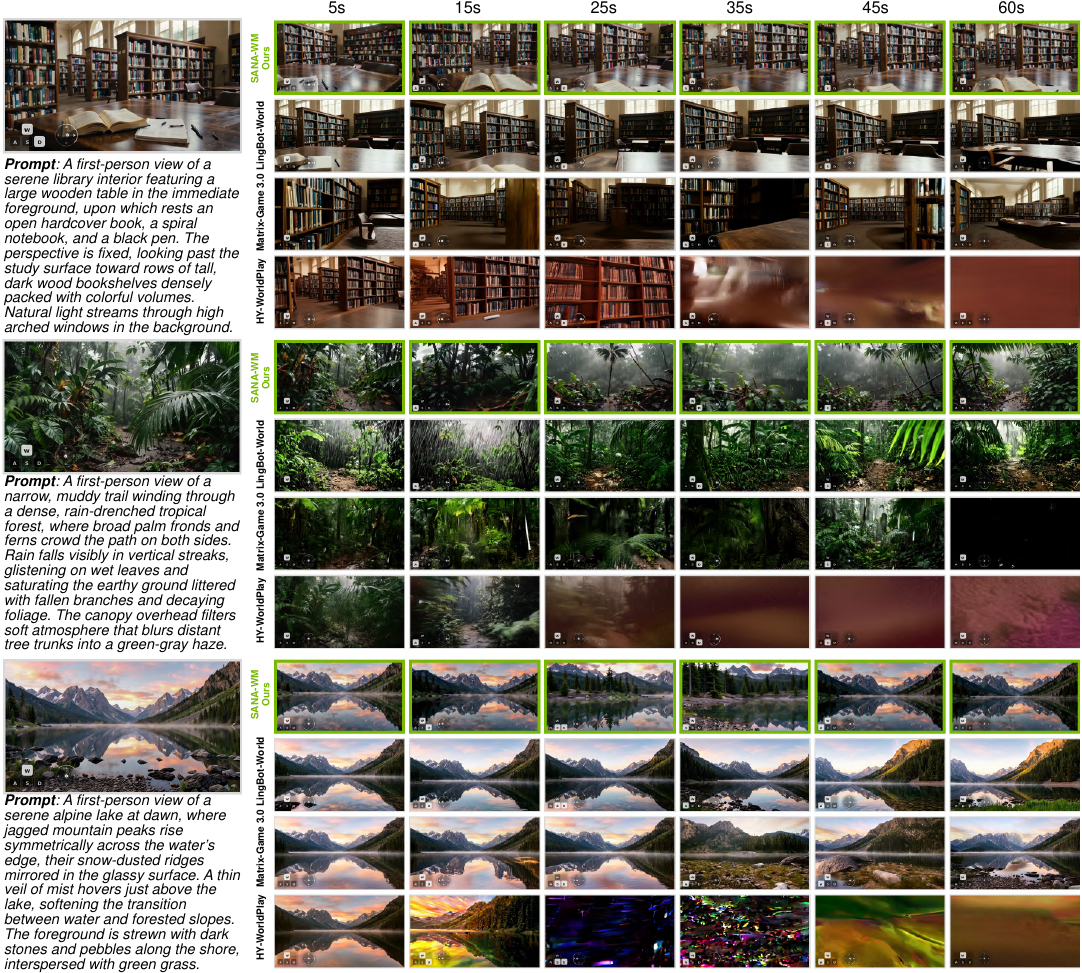}
  \caption{Qualitative comparison on 3 Hard-Trajectory 60-second videos. Green borders mark \modelname, with transparent action overlays in the bottom-left.}
  \label{fig:vis-appendix}
\end{figure}

Fig.~\ref{fig:vis-appendix} provides additional Hard-Trajectory examples beyond the main qualitative figure.
These cases emphasize viewpoint consistency under large camera motion: \modelname better preserves global layout and object identity across the minute-long rollout, whereas baselines often blur details, change scene structure, or collapse toward weak motion under the same action sequence.
The examples are selected to complement the aggregate pose, VBench, and revisit-memory results rather than to introduce a separate evaluation protocol.

\subsection{3D Reconstruction Visualization}
\label{sec:appendix_3d_recon}

\begin{figure}[t]
  \centering
  \includegraphics[width=\textwidth]{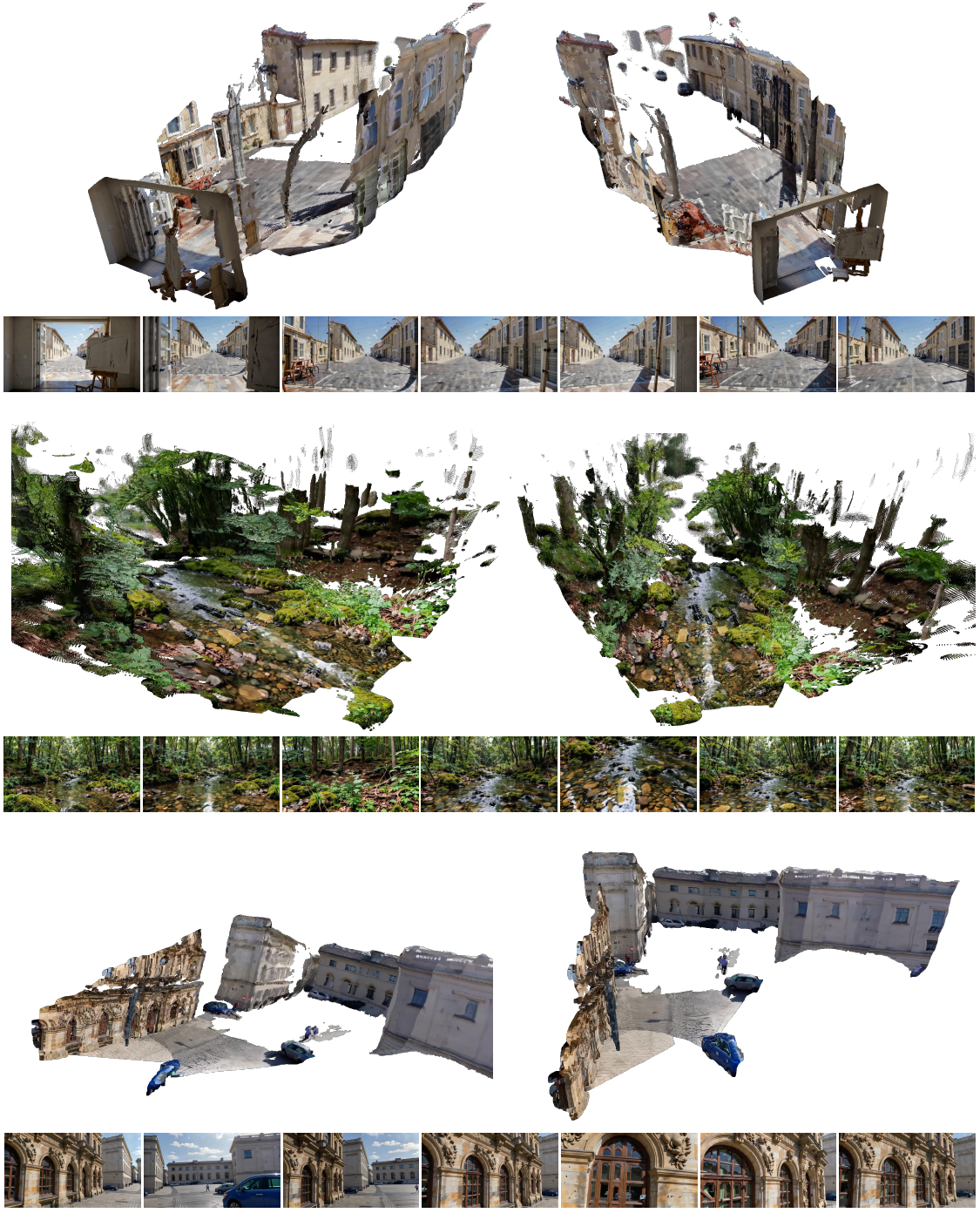}
  \caption{3D-aware qualitative demo on three benchmark videos generated by \modelname. Each example is a 60-second rollout conditioned on a Nano Banana Pro first frame; the accompanying 3D reconstruction is produced by Pi3X from the generated video.}
  \label{fig:appendix_3d_recon}
\end{figure}

Fig.~\ref{fig:appendix_3d_recon} provides an additional qualitative probe of the geometry contained in \modelname rollouts.
The three examples are benchmark results: each video is generated from a Nano Banana Pro~\citep{google2025nanobananapro} first frame and then reconstructed with Pi3X~\citep{pi3x} from the generated frames.
The recovered structures and camera paths indicate that the videos contain coherent 3D cues over minute-long camera-controlled rollouts, beyond frame-wise image quality alone.

\section{Broader Impact}
\label{sec:appendix_broader_impact}

\modelname is intended to make long-horizon world modeling more accessible for simulation, embodied AI, robotics research, and interactive content prototyping.
By reducing the training corpus size, training compute, and inference hardware needed for one-minute 720p rollouts, the system can lower the barrier for academic groups and smaller labs to study camera-controlled video generation, long-horizon scene consistency, and action-conditioned evaluation.
The same efficiency may also reduce the marginal computational cost of generating evaluation data or synthetic rollouts, although total environmental cost still depends on how widely and how repeatedly the model is used.

\noindent The main potential negative impacts are those common to high-quality video generation and simulation systems.
Generated videos may be mistaken for real observations if provenance is not documented, and simulated rollouts may be over-interpreted as faithful predictions in safety-critical settings such as robotics, autonomous driving, or physical planning.
The model also inherits biases and coverage limitations from its public video sources, generated benchmark images, and filtering/captioning tools, so performance should not be assumed to transfer uniformly across cultures, environments, rare viewpoints, or sensitive human-centered scenes.
Practical deployments should therefore document data provenance, generation settings, intended use, and known failure modes; retain watermarks or other provenance signals when available; avoid presenting generated videos as real-world evidence; and verify downstream decisions with real data and task-specific safety checks.
Tab.~\ref{tab:asset_terms} summarizes the public license or terms status of the main existing assets and tools used by our pipeline.

\section{Existing Assets and Tool Terms}
\label{sec:appendix_asset_terms}

For transparency, Tab.~\ref{tab:asset_terms} summarizes the public license or terms status we found for the main existing assets and external tools used in our data construction, annotation, evaluation, and refinement pipeline.
We cite the original projects and follow their release terms; the benchmark images and prompts are used as internal evaluation inputs rather than as a separately released dataset in this submission.

\begin{table}[t]
  \centering
  \caption{Public license or terms status for existing assets and external tools used by \modelname. ``Project terms'' indicates that the official release uses a custom or non-standard access agreement rather than a standard SPDX-style license.}
  \label{tab:asset_terms}
  \scriptsize
  \setlength{\tabcolsep}{3pt}
  \renewcommand{\arraystretch}{1.08}
  \begin{tabular}{@{}p{0.20\linewidth}p{0.27\linewidth}p{0.44\linewidth}@{}}
    \toprule
    \textbf{Asset / tool} & \textbf{Use in this work} & \textbf{Public license or terms status} \\
    \midrule
    SpatialVID-HQ~\citep{spatialvid} & Real-video training source & CC-BY-NC-SA 4.0; gated Hugging Face access requires agreement to non-commercial terms. \\
    DL3DV-10K~\citep{dl3dv} & Static 3D scenes, GT poses, and 3DGS augmentation & Custom DL3DV project terms; access requires accepting the dataset terms rather than a standard open-source license. \\
    OmniWorld~\citep{omniworld} & Synthetic/game data and held-out camera-control validation & CC-BY-NC-SA 4.0 on the public Hugging Face release. \\
    Sekai~\citep{sekai} & Game and walking-video training sources & Public project/data release; no standard license was clearly specified on the project page we checked, so use should follow the project access terms. \\
    MiraData~\citep{miradata} & Long real-video training source & Public project release; the repository lists GPL-3.0, while source videos may remain subject to their original hosting terms. \\
    ViPE~\citep{vipe} & Camera-pose annotation engine & Apache-2.0 code release, with third-party components under their respective licenses. \\
    Pi3X / Pi3~\citep{pi3x} & Pose/depth recovery and evaluation & BSD-3-Clause code; public model weights are released for non-commercial use (CC BY-NC 4.0). \\
    MoGe-2~\citep{moge2} & Metric-scale depth prior & Public Microsoft MoGe release; repository licensing includes permissive MIT/Apache-style terms, while model-weight terms should be checked from the model card. \\
    FCGS~\citep{fcgs} & 3D Gaussian Splatting reconstruction for DL3DV augmentation & Public research code; no standard license was clearly specified on the project page we checked. \\
    DiFix3D~\citep{difix3d} & Refinement of 3DGS-rendered videos & NVIDIA research release; model/code use is governed by NVIDIA terms for research and development/non-commercial use. \\
    Qwen3.5 VLM~\citep{qwen3_5} & Content filtering and scene-static captioning & Apache-2.0 public model/code release. \\
    Nano Banana Pro~\citep{google2025nanobananapro} & Generation of first-frame evaluation images & Google/Gemini product terms apply; generated images are marked with SynthID according to Google's product documentation. \\
    LTX-2 / LTX-2.3~\citep{ltx2} & LTX2 VAE and long-video refiner initialization & LTX-2 Community License Agreement for model weights, code, and related materials. \\
    \bottomrule
  \end{tabular}
\end{table}

\clearpage

{
  \small
  \bibliographystyle{unsrt}
  \bibliography{main}

\begin{thebibliography}{100}

\bibitem{world_models}
David Ha and J{\"u}rgen Schmidhuber.
\newblock World models.
\newblock {\em arXiv preprint arXiv:1803.10122}, 2(3):440, 2018.

\bibitem{genie3}
Jack Parker-Holder and Shlomi Fruchter.
\newblock Genie 3: A new frontier for world models.
\newblock \url{https://deepmind.google/en/blog/genie-3-a-new-frontier-for-world-models/}, 2025.
\newblock Google DeepMind blog post, August 2025.

\bibitem{gaia1}
Anthony Hu, Lloyd Russell, Hudson Yeo, Zak Murez, George Fedoseev, Alex Kendall, Jamie Shotton, and Gianluca Corrado.
\newblock Gaia-1: A generative world model for autonomous driving.
\newblock {\em arXiv preprint arXiv:2309.17080}, 2023.

\bibitem{dreamdojo}
Shenyuan Gao, William Liang, Kaiyuan Zheng, Ayaan Malik, Seonghyeon Ye, Sihyun Yu, Wei-Cheng Tseng, Yuzhu Dong, Kaichun Mo, Chen-Hsuan Lin, et~al.
\newblock Dreamdojo: A generalist robot world model from large-scale human videos.
\newblock {\em arXiv preprint arXiv:2602.06949}, 2026.

\bibitem{aether}
Haoyi Zhu, Yifan Wang, Jianjun Zhou, Wenzheng Chang, Yang Zhou, Zizun Li, Junyi Chen, Chunhua Shen, Jiangmiao Pang, and Tong He.
\newblock Aether: Geometric-aware unified world modeling.
\newblock In {\em Proceedings of the IEEE/CVF International Conference on Computer Vision}, pages 8535--8546, 2025.

\bibitem{hy_worldplay}
Wenqiang Sun, Haiyu Zhang, Haoyuan Wang, Junta Wu, Zehan Wang, Zhenwei Wang, Yunhong Wang, Jun Zhang, Tengfei Wang, and Chunchao Guo.
\newblock Worldplay: Towards long-term geometric consistency for real-time interactive world modeling.
\newblock {\em arXiv preprint arXiv:2512.14614}, 2025.

\bibitem{lingbot_world}
Robbyant Team, Zelin Gao, Qiuyu Wang, Yanhong Zeng, Jiapeng Zhu, Ka~Leong Cheng, Yixuan Li, Hanlin Wang, Yinghao Xu, Shuailei Ma, et~al.
\newblock Advancing open-source world models.
\newblock {\em arXiv preprint arXiv:2601.20540}, 2026.

\bibitem{infinite_world}
Ruiqi Wu, Xuanhua He, Meng Cheng, Tianyu Yang, Yong Zhang, Zhuoliang Kang, Xunliang Cai, Xiaoming Wei, Chunle Guo, Chongyi Li, et~al.
\newblock Infinite-world: Scaling interactive world models to 1000-frame horizons via pose-free hierarchical memory.
\newblock {\em arXiv preprint arXiv:2602.02393}, 2026.

\bibitem{matrixgame3}
Zile Wang, Zexiang Liu, Jaixing Li, Kaichen Huang, Baixin Xu, Fei Kang, Mengyin An, Peiyu Wang, Biao Jiang, Yichen Wei, et~al.
\newblock Matrix-game 3.0: Real-time and streaming interactive world model with long-horizon memory.
\newblock {\em arXiv preprint arXiv:2604.08995}, 2026.

\bibitem{ltx2}
Yoav HaCohen, Benny Brazowski, Nisan Chiprut, Yaki Bitterman, Andrew Kvochko, Avishai Berkowitz, Daniel Shalem, Daphna Lifschitz, Dudu Moshe, Eitan Porat, et~al.
\newblock Ltx-2: Efficient joint audio-visual foundation model.
\newblock {\em arXiv preprint arXiv:2601.03233}, 2026.

\bibitem{gated_deltanet}
Songlin Yang, Jan Kautz, and Ali Hatamizadeh.
\newblock Gated delta networks: Improving mamba2 with delta rule.
\newblock {\em arXiv preprint arXiv:2412.06464}, 2024.

\bibitem{ucpe}
Cheng Zhang, Boying Li, Meng Wei, Yan-Pei Cao, Camilo~Cruz Gambardella, Dinh Phung, and Jianfei Cai.
\newblock Unified camera positional encoding for controlled video generation.
\newblock {\em arXiv preprint arXiv:2512.07237}, 2025.

\bibitem{vipe}
Jiahui Huang, Qunjie Zhou, Hesam Rabeti, Aleksandr Korovko, Huan Ling, Xuanchi Ren, Tianchang Shen, Jun Gao, Dmitry Slepichev, Chen-Hsuan Lin, et~al.
\newblock Vipe: Video pose engine for 3d geometric perception.
\newblock {\em arXiv preprint arXiv:2508.10934}, 2025.

\bibitem{pi3x}
Yifan Wang, Jianjun Zhou, Haoyi Zhu, Wenzheng Chang, Yang Zhou, Zizun Li, Junyi Chen, Jiangmiao Pang, Chunhua Shen, and Tong He.
\newblock {$\pi^3$}: Permutation-equivariant visual geometry learning.
\newblock {\em arXiv preprint arXiv:2507.13347}, 2025.

\bibitem{moge2}
Ruicheng Wang, Sicheng Xu, Yue Dong, Yu~Deng, Jianfeng Xiang, Zelong Lv, Guangzhong Sun, Xin Tong, and Jiaolong Yang.
\newblock Moge-2: Accurate monocular geometry with metric scale and sharp details.
\newblock {\em arXiv preprint arXiv:2507.02546}, 2025.

\bibitem{google2025nanobananapro}
{Google Blog}.
\newblock Introducing {Nano Banana Pro}.
\newblock \url{https://blog.google/innovation-and-ai/products/nano-banana-pro/}, 2025.

\bibitem{stable_video_diffusion}
Andreas Blattmann, Tim Dockhorn, Sumith Kulal, Daniel Mendelevitch, Maciej Kilian, Dominik Lorenz, Yam Levi, Zion English, Vikram Voleti, Adam Letts, et~al.
\newblock Stable video diffusion: Scaling latent video diffusion models to large datasets.
\newblock {\em arXiv preprint arXiv:2311.15127}, 2023.

\bibitem{sora}
{OpenAI}.
\newblock Video generation models as world simulators.
\newblock \url{https://openai.com/index/video-generation-models-as-world-simulators/}, 2024.
\newblock Technical report, February 2024.

\bibitem{cogvideox}
Zhuoyi Yang, Jiayan Teng, Wendi Zheng, Ming Ding, Shiyu Huang, Jiazheng Xu, Yuanming Yang, Wenyi Hong, Xiaohan Zhang, Guanyu Feng, et~al.
\newblock Cogvideox: Text-to-video diffusion models with an expert transformer.
\newblock {\em arXiv preprint arXiv:2408.06072}, 2024.

\bibitem{wan}
Team Wan, Ang Wang, Baole Ai, Bin Wen, Chaojie Mao, Chen-Wei Xie, Di~Chen, Feiwu Yu, Haiming Zhao, Jianxiao Yang, et~al.
\newblock Wan: Open and advanced large-scale video generative models.
\newblock {\em arXiv preprint arXiv:2503.20314}, 2025.

\bibitem{hunyuanvideo}
Weijie Kong, Qi~Tian, Zijian Zhang, Rox Min, Zuozhuo Dai, Jin Zhou, Jiangfeng Xiong, Xin Li, Bo~Wu, Jianwei Zhang, et~al.
\newblock Hunyuanvideo: A systematic framework for large video generative models.
\newblock {\em arXiv preprint arXiv:2412.03603}, 2024.

\bibitem{moviegen}
Adam Polyak, Amit Zohar, Andrew Brown, Andros Tjandra, Animesh Sinha, Ann Lee, Apoorv Vyas, Bowen Shi, Chih-Yao Ma, Ching-Yao Chuang, et~al.
\newblock Movie gen: A cast of media foundation models.
\newblock {\em arXiv preprint arXiv:2410.13720}, 2024.

\bibitem{cosmos}
Niket Agarwal, Arslan Ali, Maciej Bala, Yogesh Balaji, Erik Barker, Tiffany Cai, Prithvijit Chattopadhyay, Yongxin Chen, Yin Cui, Yifan Ding, et~al.
\newblock Cosmos world foundation model platform for physical ai.
\newblock {\em arXiv preprint arXiv:2501.03575}, 2025.

\bibitem{ltx_video}
Yoav HaCohen, Nisan Chiprut, Benny Brazowski, Daniel Shalem, Dudu Moshe, Eitan Richardson, Eran Levin, Guy Shiran, Nir Zabari, Ori Gordon, et~al.
\newblock Ltx-video: Realtime video latent diffusion.
\newblock {\em arXiv preprint arXiv:2501.00103}, 2024.

\bibitem{chen2025sana}
Junsong Chen, Yuyang Zhao, Jincheng Yu, Ruihang Chu, Junyu Chen, Shuai Yang, Xianbang Wang, Yicheng Pan, Daquan Zhou, Huan Ling, et~al.
\newblock Sana-video: Efficient video generation with block linear diffusion transformer.
\newblock {\em arXiv preprint arXiv:2509.24695}, 2025.

\bibitem{skyreels_v2}
Guibin Chen, Dixuan Lin, Jiangping Yang, Chunze Lin, Junchen Zhu, Mingyuan Fan, Hao Zhang, Sheng Chen, Zheng Chen, Chengcheng Ma, et~al.
\newblock Skyreels-v2: Infinite-length film generative model.
\newblock {\em arXiv preprint arXiv:2504.13074}, 2025.

\bibitem{magi}
Hansi Teng, Hongyu Jia, Lei Sun, Lingzhi Li, Maolin Li, Mingqiu Tang, Shuai Han, Tianning Zhang, WQ~Zhang, Weifeng Luo, et~al.
\newblock Magi-1: Autoregressive video generation at scale.
\newblock {\em arXiv preprint arXiv:2505.13211}, 2025.

\bibitem{huang2024selfforcing}
Xun Huang, Zhengqi Li, Guande He, Mingyuan Zhou, and Eli Shechtman.
\newblock Self forcing: Bridging the train-test gap in autoregressive video diffusion.
\newblock {\em arXiv preprint arXiv:2506.08009}, 2025.

\bibitem{longlive}
Shuai Yang, Wei Huang, Ruihang Chu, Yicheng Xiao, Yuyang Zhao, Xianbang Wang, Muyang Li, Enze Xie, Yingcong Chen, Yao Lu, et~al.
\newblock Longlive: Real-time interactive long video generation.
\newblock {\em arXiv preprint arXiv:2509.22622}, 2025.

\bibitem{dreamerv3}
Danijar Hafner, Jurgis Pasukonis, Jimmy Ba, and Timothy Lillicrap.
\newblock Mastering diverse domains through world models.
\newblock {\em arXiv preprint arXiv:2301.04104}, 2023.

\bibitem{i_jepa}
Mahmoud Assran, Quentin Duval, Ishan Misra, Piotr Bojanowski, Pascal Vincent, Michael Rabbat, Yann LeCun, and Nicolas Ballas.
\newblock Self-supervised learning from images with a joint-embedding predictive architecture.
\newblock In {\em Proceedings of the IEEE/CVF conference on computer vision and pattern recognition}, pages 15619--15629, 2023.

\bibitem{v_jepa}
Adrien Bardes, Quentin Garrido, Jean Ponce, Xinlei Chen, Michael Rabbat, Yann LeCun, Mahmoud Assran, and Nicolas Ballas.
\newblock Revisiting feature prediction for learning visual representations from video, 2024.

\bibitem{v_jepa2}
Mido Assran, Adrien Bardes, David Fan, Quentin Garrido, Russell Howes, Matthew Muckley, Ammar Rizvi, Claire Roberts, Koustuv Sinha, Artem Zholus, et~al.
\newblock V-jepa 2: Self-supervised video models enable understanding, prediction and planning.
\newblock {\em arXiv preprint arXiv:2506.09985}, 2025.

\bibitem{genie}
Jake Bruce, Michael~D Dennis, Ashley Edwards, Jack Parker-Holder, Yuge Shi, Edward Hughes, Matthew Lai, Aditi Mavalankar, Richie Steigerwald, Chris Apps, et~al.
\newblock Genie: Generative interactive environments.
\newblock In {\em Forty-first International Conference on Machine Learning}, 2024.

\bibitem{genie2}
{Google DeepMind}.
\newblock Genie 2: A large-scale foundation world model.
\newblock \url{https://deepmind.google/blog/genie-2-a-large-scale-foundation-world-model/}, 2024.
\newblock Blog post.

\bibitem{gamengen}
Dani Valevski, Yaniv Leviathan, Moab Arar, and Shlomi Fruchter.
\newblock Diffusion models are real-time game engines.
\newblock {\em arXiv preprint arXiv:2408.14837}, 2024.

\bibitem{oasis}
Decart and Etched.
\newblock Oasis: A universe in a transformer.
\newblock \url{https://oasis-model.github.io/}, 2024.
\newblock Technical report.

\bibitem{gamegen}
Haoxuan Che, Xuanhua He, Quande Liu, Cheng Jin, and Hao Chen.
\newblock Gamegen-x: Interactive open-world game video generation.
\newblock {\em arXiv preprint arXiv:2411.00769}, 2024.

\bibitem{matrixgame}
Yifan Zhang, Chunli Peng, Boyang Wang, Puyi Wang, Qingcheng Zhu, Fei Kang, Biao Jiang, Zedong Gao, Eric Li, Yang Liu, et~al.
\newblock Matrix-game: Interactive world foundation model.
\newblock {\em arXiv preprint arXiv:2506.18701}, 2025.

\bibitem{relic}
Yicong Hong, Yiqun Mei, Chongjian Ge, Yiran Xu, Yang Zhou, Sai Bi, Yannick Hold-Geoffroy, Mike Roberts, Matthew Fisher, Eli Shechtman, et~al.
\newblock Relic: Interactive video world model with long-horizon memory.
\newblock {\em arXiv preprint arXiv:2512.04040}, 2025.

\bibitem{live_world}
Junchao Huang, Ziyang Ye, Xinting Hu, Tianyu He, Guiyu Zhang, Shaoshuai Shi, Jiang Bian, and Li~Jiang.
\newblock Live: Long-horizon interactive video world modeling.
\newblock {\em arXiv preprint arXiv:2602.03747}, 2026.

\bibitem{astra}
Yixuan Zhu, Jiaqi Feng, Wenzhao Zheng, Yuan Gao, Xin Tao, Pengfei Wan, Jie Zhou, and Jiwen Lu.
\newblock Astra: General interactive world model with autoregressive denoising.
\newblock {\em arXiv preprint arXiv:2512.08931}, 2025.

\bibitem{magicworld}
Guangyuan Li, Siming Zheng, Shuolin Xu, Jinwei Chen, Bo~Li, Xiaobin Hu, Lei Zhao, and Peng-Tao Jiang.
\newblock Magicworld: Interactive geometry-driven video world exploration.
\newblock {\em arXiv preprint arXiv:2511.18886}, 2025.

\bibitem{yume}
Xiaofeng Mao, Zhen Li, Chuanhao Li, Xiaojie Xu, Kaining Ying, Tong He, Jiangmiao Pang, Yu~Qiao, and Kaipeng Zhang.
\newblock Yume-1.5: A text-controlled interactive world generation model.
\newblock {\em arXiv preprint arXiv:2512.22096}, 2025.

\bibitem{dreamzero}
Seonghyeon Ye, Yunhao Ge, Kaiyuan Zheng, Shenyuan Gao, Sihyun Yu, George Kurian, Suneel Indupuru, You~Liang Tan, Chuning Zhu, Jiannan Xiang, et~al.
\newblock World action models are zero-shot policies.
\newblock {\em arXiv preprint arXiv:2602.15922}, 2026.

\bibitem{unisim}
Sherry Yang, Yilun Du, Kamyar Ghasemipour, Jonathan Tompson, Leslie Kaelbling, Dale Schuurmans, and Pieter Abbeel.
\newblock Learning interactive real-world simulators.
\newblock {\em arXiv preprint arXiv:2310.06114}, 2023.

\bibitem{worldcam}
Jisu Nam, Yicong Hong, Chun-Hao~Paul Huang, Feng Liu, JoungBin Lee, Jiyoung Kim, Siyoon Jin, Yunsung Lee, Jaeyoon Jung, Suhwan Choi, et~al.
\newblock Worldcam: Interactive autoregressive 3d gaming worlds with camera pose as a unifying geometric representation.
\newblock {\em arXiv preprint arXiv:2603.16871}, 2026.

\bibitem{ucm}
Tianxing Xu, Zixuan Wang, Guangyuan Wang, Li~Hu, Zhongyi Zhang, Peng Zhang, Bang Zhang, and Song-Hai Zhang.
\newblock Ucm: Unifying camera control and memory with time-aware positional encoding warping for world models.
\newblock {\em arXiv preprint arXiv:2602.22960}, 2026.

\bibitem{captain_safari}
Yu-Cheng Chou, Xingrui Wang, Yitong Li, Jiahao Wang, Hanting Liu, Cihang Xie, Alan Yuille, and Junfei Xiao.
\newblock Captain safari: A world engine with pose-aligned 3d memory, 2026.

\bibitem{deepverse}
Junyi Chen, Haoyi Zhu, Xianglong He, Yifan Wang, Jianjun Zhou, Wenzheng Chang, Yang Zhou, Zizun Li, Zhoujie Fu, Jiangmiao Pang, et~al.
\newblock Deepverse: 4d autoregressive video generation as a world model.
\newblock {\em arXiv preprint arXiv:2506.01103}, 2025.

\bibitem{drivedreamer}
Xiaofeng Wang, Zheng Zhu, Guan Huang, Xinze Chen, Jiagang Zhu, and Jiwen Lu.
\newblock Drivedreamer: towards real-world-driven world models for autonomous driving.
\newblock {\em arXiv preprint arXiv:2309.09777}, 2023.

\bibitem{driveworld}
Chen Min, Dawei Zhao, Liang Xiao, Jian Zhao, Xinli Xu, Zheng Zhu, Lei Jin, Jianshu Li, Yulan Guo, Junliang Xing, et~al.
\newblock Driveworld: 4d pre-trained scene understanding via world models for autonomous driving.
\newblock In {\em Proceedings of the IEEE/CVF conference on computer vision and pattern recognition}, pages 15522--15533, 2024.

\bibitem{mosaicmem}
Wei Yu, Runjia Qian, Yumeng Li, Liquan Wang, Songheng Yin, Dennis Anthony, Yang Ye, Yidi Li, Weiwei Wan, Animesh Garg, et~al.
\newblock Mosaicmem: Hybrid spatial memory for controllable video world models.
\newblock {\em arXiv preprint arXiv:2603.17117}, 2026.

\bibitem{memorize_when_needed}
Yanjun Guo, Zhengqiang Zhang, Pengfei Wang, Xinyue Liang, Zhiyuan Ma, and Lei Zhang.
\newblock Memorize when needed: Decoupled memory control for spatially consistent long-horizon video generation, 2026.

\bibitem{vggt_world}
Xiangyu Sun, Shijie Wang, Fengyi Zhang, Lin Liu, Caiyan Jia, Ziying Song, Zi~Huang, and Yadan Luo.
\newblock Vggt-world: Transforming vggt into an autoregressive geometry world model.
\newblock {\em arXiv preprint arXiv:2603.12655}, 2026.

\bibitem{versecrafter}
Sixiao Zheng, Minghao Yin, Wenbo Hu, Xiaoyu Li, Ying Shan, and Yanwei Fu.
\newblock Versecrafter: Dynamic realistic video world model with 4d geometric control.
\newblock {\em arXiv preprint arXiv:2601.05138}, 2026.

\bibitem{hy_world2}
Team HY-World, Chenjie Cao, Xuhui Zuo, Zhenwei Wang, Yisu Zhang, Junta Wu, Zhenyang Liu, Yuning Gong, Yang Liu, Bo~Yuan, et~al.
\newblock Hy-world 2.0: A multi-modal world model for reconstructing, generating, and simulating 3d worlds.
\newblock {\em arXiv preprint arXiv:2604.14268}, 2026.

\bibitem{inspatio_world}
InSpatio Team, Donghui Shen, Guofeng Zhang, Haomin Liu, Haoyu Ji, Hujun Bao, Hongjia Zhai, Jialin Liu, Jing Guo, Nan Wang, et~al.
\newblock Inspatio-world: A real-time 4d world simulator via spatiotemporal autoregressive modeling.
\newblock {\em arXiv preprint arXiv:2604.07209}, 2026.

\bibitem{cameractrl}
Hao He, Yinghao Xu, Yuwei Guo, Gordon Wetzstein, Bo~Dai, Hongsheng Li, and Ceyuan Yang.
\newblock Cameractrl: Enabling camera control for text-to-video generation.
\newblock {\em arXiv preprint arXiv:2404.02101}, 2024.

\bibitem{motionctrl}
Zhouxia Wang, Ziyang Yuan, Xintao Wang, Yaowei Li, Tianshui Chen, Menghan Xia, Ping Luo, and Ying Shan.
\newblock Motionctrl: A unified and flexible motion controller for video generation.
\newblock In {\em ACM SIGGRAPH 2024 Conference Papers}, pages 1--11, 2024.

\bibitem{camco}
Dejia Xu, Weili Nie, Chao Liu, Sifei Liu, Jan Kautz, Zhangyang Wang, and Arash Vahdat.
\newblock Camco: Camera-controllable 3d-consistent image-to-video generation.
\newblock {\em arXiv preprint arXiv:2406.02509}, 2024.

\bibitem{viewcrafter}
Wangbo Yu, Jinbo Xing, Li~Yuan, Wenbo Hu, Xiaoyu Li, Zhipeng Huang, Xiangjun Gao, Tien-Tsin Wong, Ying Shan, and Yonghong Tian.
\newblock Viewcrafter: Taming video diffusion models for high-fidelity novel view synthesis.
\newblock {\em arXiv preprint arXiv:2409.02048}, 2024.

\bibitem{seva}
Jensen Zhou, Hang Gao, Vikram Voleti, Aaryaman Vasishta, Chun-Han Yao, Mark Boss, Philip Torr, Christian Rupprecht, and Varun Jampani.
\newblock Stable virtual camera: Generative view synthesis with diffusion models.
\newblock In {\em Proceedings of the IEEE/CVF International Conference on Computer Vision}, pages 12405--12414, 2025.

\bibitem{plucker_lfn}
Vincent Sitzmann, Semon Rezchikov, Bill Freeman, Josh Tenenbaum, and Fredo Durand.
\newblock Light field networks: Neural scene representations with single-evaluation rendering.
\newblock {\em Advances in Neural Information Processing Systems}, 34:19313--19325, 2021.

\bibitem{li2025cameras}
Ruilong Li, Brent Yi, Junchen Liu, Hang Gao, Yi~Ma, and Angjoo Kanazawa.
\newblock Cameras as relative positional encoding.
\newblock {\em arXiv preprint arXiv:2507.10496}, 2025.

\bibitem{vggt}
Jianyuan Wang, Minghao Chen, Nikita Karaev, Andrea Vedaldi, Christian Rupprecht, and David Novotny.
\newblock Vggt: Visual geometry grounded transformer.
\newblock In {\em Proceedings of the Computer Vision and Pattern Recognition Conference}, pages 5294--5306, 2025.

\bibitem{wint3r}
Zizun Li, Jianjun Zhou, Yifan Wang, Haoyu Guo, Wenzheng Chang, Yang Zhou, Haoyi Zhu, Junyi Chen, Chunhua Shen, and Tong He.
\newblock Wint3r: Window-based streaming reconstruction with camera token pool.
\newblock {\em arXiv preprint arXiv:2509.05296}, 2025.

\bibitem{flash_attention2}
Tri Dao.
\newblock Flashattention-2: Faster attention with better parallelism and work partitioning.
\newblock {\em arXiv preprint arXiv:2307.08691}, 2023.

\bibitem{linear_attention}
Angelos Katharopoulos, Apoorv Vyas, Nikolaos Pappas, and Fran{\c{c}}ois Fleuret.
\newblock Transformers are rnns: Fast autoregressive transformers with linear attention.
\newblock In {\em International conference on machine learning}, pages 5156--5165. PMLR, 2020.

\bibitem{performer}
Krzysztof Choromanski, Valerii Likhosherstov, David Dohan, Xingyou Song, Andreea Gane, Tamas Sarlos, Peter Hawkins, Jared Davis, Afroz Mohiuddin, Lukasz Kaiser, et~al.
\newblock Rethinking attention with performers.
\newblock {\em arXiv preprint arXiv:2009.14794}, 2020.

\bibitem{gla}
Songlin Yang, Bailin Wang, Yikang Shen, Rameswar Panda, and Yoon Kim.
\newblock Gated linear attention transformers with hardware-efficient training.
\newblock {\em arXiv preprint arXiv:2312.06635}, 2023.

\bibitem{rwkv}
Bo~Peng, Eric Alcaide, Quentin Anthony, Alon Albalak, Samuel Arcadinho, Stella Biderman, Huanqi Cao, Xin Cheng, Michael Chung, Leon Derczynski, et~al.
\newblock Rwkv: Reinventing rnns for the transformer era.
\newblock In {\em Findings of the association for computational linguistics: EMNLP 2023}, pages 14048--14077, 2023.

\bibitem{retnet}
Yutao Sun, Li~Dong, Shaohan Huang, Shuming Ma, Yuqing Xia, Jilong Xue, Jianyong Wang, and Furu Wei.
\newblock Retentive network: A successor to transformer for large language models.
\newblock {\em arXiv preprint arXiv:2307.08621}, 2023.

\bibitem{hyena}
Michael Poli, Stefano Massaroli, Eric Nguyen, Daniel~Y Fu, Tri Dao, Stephen Baccus, Yoshua Bengio, Stefano Ermon, and Christopher R{\'e}.
\newblock Hyena hierarchy: Towards larger convolutional language models.
\newblock In {\em International Conference on Machine Learning}, pages 28043--28078. PMLR, 2023.

\bibitem{mamba}
Albert Gu and Tri Dao.
\newblock Mamba: Linear-time sequence modeling with selective state spaces.
\newblock {\em arXiv preprint arXiv:2312.00752}, 2023.

\bibitem{mamba2}
Tri Dao and Albert Gu.
\newblock Transformers are ssms: Generalized models and efficient algorithms through structured state space duality.
\newblock {\em arXiv preprint arXiv:2405.21060}, 2024.

\bibitem{ttt_layers}
Yu~Sun, Xinhao Li, Karan Dalal, Jiarui Xu, Arjun Vikram, Genghan Zhang, Yann Dubois, Xinlei Chen, Xiaolong Wang, Sanmi Koyejo, et~al.
\newblock Learning to (learn at test time): Rnns with expressive hidden states.
\newblock {\em arXiv preprint arXiv:2407.04620}, 2024.

\bibitem{qwen3}
An~Yang, Anfeng Li, Baosong Yang, Beichen Zhang, Binyuan Hui, Bo~Zheng, Bowen Yu, Chang Gao, Chengen Huang, Chenxu Lv, et~al.
\newblock Qwen3 technical report.
\newblock {\em arXiv preprint arXiv:2505.09388}, 2025.

\bibitem{qwen3_next}
{Qwen Team}.
\newblock Qwen3-next: Hybrid attention with gated deltanet.
\newblock \url{https://huggingface.co/collections/Qwen/qwen3-next}, 2025.
\newblock Model collections.

\bibitem{kimi_linear}
Kimi Team, Yu~Zhang, Zongyu Lin, Xingcheng Yao, Jiaxi Hu, Fanqing Meng, Chengyin Liu, Xin Men, Songlin Yang, Zhiyuan Li, et~al.
\newblock Kimi linear: An expressive, efficient attention architecture.
\newblock {\em arXiv preprint arXiv:2510.26692}, 2025.

\bibitem{kimi_k2}
Kimi Team, Yifan Bai, Yiping Bao, Y~Charles, Cheng Chen, Guanduo Chen, Haiting Chen, Huarong Chen, Jiahao Chen, Ningxin Chen, et~al.
\newblock Kimi k2: Open agentic intelligence.
\newblock {\em arXiv preprint arXiv:2507.20534}, 2025.

\bibitem{sana}
Enze Xie, Junsong Chen, Junyu Chen, Han Cai, Haotian Tang, Yujun Lin, Zhekai Zhang, Muyang Li, Ligeng Zhu, Yao Lu, et~al.
\newblock Sana: Efficient high-resolution image synthesis with linear diffusion transformers.
\newblock {\em arXiv preprint arXiv:2410.10629}, 2024.

\bibitem{dc_ae}
Junyu Chen, Han Cai, Junsong Chen, Enze Xie, Shang Yang, Haotian Tang, Muyang Li, Yao Lu, and Song Han.
\newblock Deep compression autoencoder for efficient high-resolution diffusion models.
\newblock {\em arXiv preprint arXiv:2410.10733}, 2024.

\bibitem{dc-video-gen}
Junyu Chen, Wenkun He, Yuchao Gu, Yuyang Zhao, Jincheng Yu, Junsong Chen, Dongyun Zou, Yujun Lin, Zhekai Zhang, Muyang Li, Haocheng Xi, Ligeng Zhu, Enze Xie, Song Han, and Han Cai.
\newblock Dc-videogen: Efficient video generation with deep compression video autoencoder, 2025.

\bibitem{realestate10k}
Tinghui Zhou, Richard Tucker, John Flynn, Graham Fyffe, and Noah Snavely.
\newblock Stereo magnification: Learning view synthesis using multiplane images.
\newblock {\em arXiv preprint arXiv:1805.09817}, 2018.

\bibitem{dl3dv}
Lu~Ling, Yichen Sheng, Zhi Tu, Wentian Zhao, Cheng Xin, Kun Wan, Lantao Yu, Qianyu Guo, Zixun Yu, Yawen Lu, et~al.
\newblock Dl3dv-10k: A large-scale scene dataset for deep learning-based 3d vision.
\newblock In {\em Proceedings of the IEEE/CVF Conference on Computer Vision and Pattern Recognition}, pages 22160--22169, 2024.

\bibitem{spatialvid}
Jiahao Wang, Yufeng Yuan, Rujie Zheng, Youtian Lin, Jian Gao, Lin-Zhuo Chen, Yajie Bao, Yi~Zhang, Chang Zeng, Yanxi Zhou, et~al.
\newblock Spatialvid: A large-scale video dataset with spatial annotations.
\newblock {\em arXiv preprint arXiv:2509.09676}, 2025.

\bibitem{miradata}
Xuan Ju, Yiming Gao, Zhaoyang Zhang, Ziyang Yuan, Xintao Wang, Ailing Zeng, Yu~Xiong, Qiang Xu, and Ying Shan.
\newblock Miradata: A large-scale video dataset with long durations and structured captions.
\newblock {\em Advances in Neural Information Processing Systems}, 37:48955--48970, 2024.

\bibitem{omniworld}
Yang Zhou, Yifan Wang, Jianjun Zhou, Wenzheng Chang, Haoyu Guo, Zizun Li, Kaijing Ma, Xinyue Li, Yating Wang, Haoyi Zhu, et~al.
\newblock Omniworld: A multi-domain and multi-modal dataset for 4d world modeling.
\newblock {\em arXiv preprint arXiv:2509.12201}, 2025.

\bibitem{sekai}
Zhen Li, Chuanhao Li, Xiaofeng Mao, Shaoheng Lin, Ming Li, Shitian Zhao, Zhaopan Xu, Xinyue Li, Yukang Feng, Jianwen Sun, et~al.
\newblock Sekai: A video dataset towards world exploration.
\newblock {\em arXiv preprint arXiv:2506.15675}, 2025.

\bibitem{transnetv2}
Tom{\'a}s Soucek and Jakub Lokoc.
\newblock Transnet v2: An effective deep network architecture for fast shot transition detection.
\newblock In {\em Proceedings of the 32nd ACM International Conference on Multimedia}, pages 11218--11221, 2024.

\bibitem{dover}
Haoning Wu, Erli Zhang, Liang Liao, Chaofeng Chen, Jingwen Hou, Annan Wang, Wenxiu Sun, Qiong Yan, and Weisi Lin.
\newblock Exploring video quality assessment on user generated contents from aesthetic and technical perspectives.
\newblock In {\em Proceedings of the IEEE/CVF international conference on computer vision}, pages 20144--20154, 2023.

\bibitem{unimatch}
Haofei Xu, Jing Zhang, Jianfei Cai, Hamid Rezatofighi, Fisher Yu, Dacheng Tao, and Andreas Geiger.
\newblock Unifying flow, stereo and depth estimation.
\newblock {\em IEEE Transactions on Pattern Analysis and Machine Intelligence}, 45(11):13941--13958, 2023.

\bibitem{fcgs}
Yihang Chen, Qianyi Wu, Mengyao Li, Weiyao Lin, Mehrtash Harandi, and Jianfei Cai.
\newblock Fast feedforward 3d gaussian splatting compression.
\newblock {\em arXiv preprint arXiv:2410.08017}, 2024.

\bibitem{difix3d}
Jay~Zhangjie Wu, Yuxuan Zhang, Haithem Turki, Xuanchi Ren, Jun Gao, Mike~Zheng Shou, Sanja Fidler, Zan Gojcic, and Huan Ling.
\newblock Difix3d+: Improving 3d reconstructions with single-step diffusion models.
\newblock In {\em Proceedings of the IEEE/CVF Conference on Computer Vision and Pattern Recognition}, pages 26024--26035, 2025.

\bibitem{vbench}
Ziqi Huang, Yinan He, Jiashuo Yu, Fan Zhang, Chenyang Si, Yuming Jiang, Yuanhan Zhang, Tianxing Wu, Qingyang Jin, Nattapol Chanpaisit, et~al.
\newblock Vbench: Comprehensive benchmark suite for video generative models.
\newblock In {\em Proceedings of the IEEE/CVF Conference on Computer Vision and Pattern Recognition}, pages 21807--21818, 2024.

\bibitem{lpips}
Richard Zhang, Phillip Isola, Alexei~A Efros, Eli Shechtman, and Oliver Wang.
\newblock The unreasonable effectiveness of deep features as a perceptual metric.
\newblock In {\em Proceedings of the IEEE conference on computer vision and pattern recognition}, pages 586--595, 2018.

\bibitem{unterthiner2018towards}
Thomas Unterthiner, Sjoerd van Steenkiste, Karol Kurach, Raphael Marinier, Marcin Michalski, and Sylvain Gelly.
\newblock Towards accurate generative models of video: A new metric \& challenges.
\newblock {\em arXiv preprint arXiv:1812.01717}, 2018.

\bibitem{umeyama}
Shinji Umeyama.
\newblock Least-squares estimation of transformation parameters between two point patterns.
\newblock {\em IEEE Transactions on pattern analysis and machine intelligence}, 13(4):376--380, 2002.

\bibitem{dit}
William Peebles and Saining Xie.
\newblock Scalable diffusion models with transformers.
\newblock In {\em Proceedings of the IEEE/CVF international conference on computer vision}, pages 4195--4205, 2023.

\bibitem{flow_matching}
Yaron Lipman, Ricky~TQ Chen, Heli Ben-Hamu, Maximilian Nickel, and Matt Le.
\newblock Flow matching for generative modeling.
\newblock {\em arXiv preprint arXiv:2210.02747}, 2022.

\bibitem{qwen3_5}
{Qwen Team}.
\newblock {Qwen3.5}: Towards native multimodal agents, February 2026.

\end{thebibliography}
}

\end{document}